\pdfoutput=1

\documentclass[11pt]{article}

\usepackage[]{acl}

\usepackage{times}
\usepackage{enumitem}
\usepackage{latexsym}
\usepackage{todonotes}
\usepackage{floatrow}
\newfloatcommand{capbtabbox}{table}[][\FBwidth]
\usepackage{ragged2e}
\usepackage{arydshln}
\RequirePackage{xcolor}
\definecolor{darkgreen}{RGB}{0,119,0} 

\usepackage[T1]{fontenc}

\usepackage[utf8]{inputenc}
\usepackage{amsfonts}
\usepackage{amssymb}
\usepackage{bbm}

\usepackage{url}
\newcommand{\olddata}{\textsf{Harm-P}}
\newcommand{\extharmp}{\textsf{Ext-Harm-P}}
\newcommand{\content}{\textsf{CE}}
\newcommand{\embharm}{\textsf{EH}}
\newcommand{\contimg}{\textsf{CI}}
\newcommand{\conttxt}{\textsf{CT}}
\newcommand{\contmm}{\textsf{CMM}}
\newcommand{\mmlrbp}{\textsf{MMLRBP}}

\usepackage[utf8]{inputenc}
\usepackage{blindtext}
\usepackage{subcaption}
\usepackage{multirow}
\usepackage{adjustbox}
\usepackage{graphicx}
\usepackage{soul}
\usepackage{enumitem}
\usepackage{arydshln}
\usepackage[font=small,labelfont=bf]{caption}
\usepackage[multiple]{footmisc}
\newcommand{\model}{\texttt{DISARM}}
\title{\model: Detecting the Victims Targeted by Harmful Memes}

\author{Shivam Sharma$^{1,3}$, Md. Shad Akhtar$^1$, Preslav Nakov$^2$, Tanmoy Chakraborty$^1$\\
  $^1$Indraprastha Institute of Information Technology - Delhi, India  \\
  $^2$Qatar Computing Research Institute, HBKU, Doha, Qatar \\
  $^3$Wipro AI Labs, India\\
  \small\texttt{\{shivams, shad.akhtar, tanmoy\}@iiitd.ac.in}\\\small\texttt{pnakov@hbku.edu.qa}}

\begin{document}
\maketitle
\begin{abstract}
Internet memes have emerged as an increasingly popular means of communication on the Web. Although typically intended to elicit humour, they have been increasingly used to spread hatred, trolling, and cyberbullying, as well as to target specific individuals, communities, or society on political, socio-cultural, and psychological grounds. While previous work has focused on detecting harmful, hateful, and offensive memes, identifying whom they attack remains a challenging and underexplored area. Here we aim to bridge this gap. In particular, we create a dataset where we annotate each meme with its victim(s) such as the name of the targeted person(s), organization(s), and community(ies). We then propose \model\ (Detecting vIctimS targeted by hARmful Memes), a framework that uses named entity recognition and person identification to detect all entities a meme is referring to, and then, incorporates a novel contextualized multimodal deep neural network to classify whether the meme intends to harm these entities. We perform several systematic experiments on three test setups, corresponding to entities that are (a)~all seen while training, (b)~not seen as a harmful target on training, and (c)~not seen at all on training. The evaluation results show that \model\ significantly outperforms ten unimodal and multimodal systems. Finally, we show that \model\ is interpretable and comparatively more generalizable and that it can reduce the relative error rate for harmful target identification by up to 9 points absolute over several strong multimodal rivals.
\end{abstract}

\section{Introduction}

Social media offer the freedom and the means to express deeply ingrained sentiments, 
which can be done using diverse and multimodal content such as memes. Besides being popularly used to express benign humour, Internet memes have also been misused to incite extreme reactions, hatred, and to spread disinformation on a massive scale.

Numerous recent efforts have attempted to characterize harmfulness \cite{pramanick-etal-2021-momenta-multimodal}, hate speech \cite{kiela2020hateful}, and offensiveness \cite{suryawanshi-etal-2020-multimodal} within memes. Most of these efforts have been directed towards detecting malicious influence within memes, but there has been little work on identifying \textit{whom the memes target}. Besides detecting whether a meme is harmful, it is often important to know whether the meme contains an entity that is particularly targeted in a harmful way. 
This is the task we are addressing here: detecting the entities that a meme targets in a harmful way.

\begin{figure}[t!]
\centering
\subfloat[{\centering Harmful reference}\label{fig:meme_harmful}]{
\includegraphics[width=0.58\columnwidth]{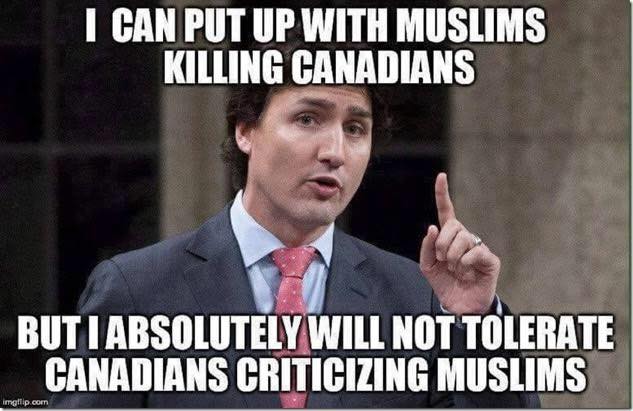}}\hspace{0.1mm}
\subfloat[{Harmless reference}\label{fig:meme_harmless}]{
\includegraphics[width=0.38\columnwidth]{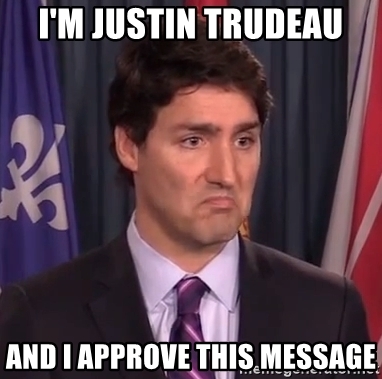}}
\caption{
(a) A meme that targets Justin Trudeau in a \emph{harmful} way, with a communal angle. (b) A \emph{non-harmful} mention of Justin Trudeau, as a benign humor.}
\label{fig:examples_harmfulness}
\end{figure}

Harmful targeting in memes is often done using satire, sarcasm, or humour in an explicit or an implicit way, aiming at attacking an individual, an organization, a community, or society in general. For example, Fig.~\ref{fig:meme_harmful} depicts Justin Trudeau, the Prime Minister of Canada, as \textit{communally biased against} Canadians, while favoring alleged \textit{killings by} Muslims, whereas Fig.~\ref{fig:meme_harmless} shows an arguably benign meme of the same person expressing subtle humour. 
Essentially, the meme in Fig.~\ref{fig:meme_harmful} \textit{harmfully} targets \emph{Justin Trudeau} directly, while causing indirect harm to \emph{Canadians} and to \emph{Muslims} as well. Note that in many cases interpreting memes and their harmful intent requires some additional background knowledge for the meme to be understood properly.

Hence, an automated system for detecting the entities targeted by harmful memes faces two major challenges:
(\emph{i})~insufficient \textit{background context}, (\emph{ii})~complexity posed by the \textit{implicit} harm, and (\emph{iii})~keyword \textit{bias} in a supervised setting.

To address these challenges, here we aim to address the task of harmful target detection in memes by formulating it as an open-ended task, where a meme can target an entity not seen on training. An end-to-end solution requires (\emph{i})~identifying the entities referred to in the meme, and (\emph{ii})~deciding whether each of these entities is being targeted in a harmful way. To address these two tasks, we perform systematic contextualization of the multimodal information presented within the meme by first performing intra-modal fusion between an external knowledge-based \textit{contextualized-entity} and the \textit{textually-embedded harmfulness} in the meme, which is followed by cross-modal fusion of the contextualized textual and visual modalities using low-rank bi-linear pooling, resulting in an enriched multimodal representation. We evaluate our model using three-level stress-testing to better assess its generalizability to unseen targets.

We create a dataset, and we propose an experimental setup and a model to address the aforementioned requirements, making the following contributions:\footnote{The source code and the dataset can be found here \url{https://github.com/LCS2-IIITD/DISARM}.}:
\begin{enumerate}
    \item We introduce the novel task of detecting the entities targeted by  harmful memes.
    \item We create a new dataset for this new task, \extharmp, by extending \olddata\ \cite{pramanick-etal-2021-momenta-multimodal} via re-annotating each harmful meme with the entity it targets.
    \item We propose \model, a novel multimodal neural architecture that uses an expressive contextualized representation for detecting harmful targeting in memes.
    \item We empirically showcase that \model\ outperforms ten unimodal and multimodal models by several points absolute in terms of macro-F1 scores in three different evaluation setups.
    \item Finally, we discuss \model's generalizability and interpretability.
\end{enumerate}

\section{Related Work}

\paragraph{Misconduct on Social Media.}

The rise in misconduct on social media is a prominent research topic. Some forms of online misconduct include rumours \cite{zhou-etal-2019-early}, fake news \cite{ALDWAIRI2018215, Shufakenews2017,CIKM2020:FANG}, misinformation \cite{gomesMisinfo2021,Claim:retrieval:context:2022}, disinformation \cite{stefano2021multidis,Survey:2022:Stance:Disinformation}, hate speech \cite{MacAvaney2019Hate, zhang2018hate,zampieri-etal-2020-semeval}, trolling \cite{cook2018troll}, and cyber-bullying \cite{Kowalski2014bullying, choudhury21cyberbully}. Some notable work in this direction includes stance \cite{stance2020yates} and rumour veracity prediction, in a multi-task learning framework \cite{kumar-carley-2019-tree}, wherein the authors proposed a Tree LSTM for characterizing online conversations. \citet{wufakenews2018} explored user and social network representations for classifying a message as genuine vs. fake. 
\citet{chengtroll2017} studied user's mood along with the online contextual discourse and demonstrated that it helps for trolling behaviour prediction on top of user's behavioural history.  
\citet{Relia_Li_Cook_Chunara_2019} studied the synergy between discrimination based on race, ethnicity, and national origin in the physical and in the virtual space.

\paragraph{Studies Focusing on Memes.}

Recent efforts have shown interest in incorporating additional contextual information for meme analysis. \citet{9582340} proposed knowledge-enriched graph neural networks that use common-sense knowledge for offensive memes detection. \citet{pramanick-etal-2021-detecting} focused on detecting COVID-19-related harmful memes and highlighted the challenge posed by the inherent biases within the existing multimodal systems.  \citet{pramanick-etal-2021-momenta-multimodal} released another dataset focusing on US Politics and proposed a multimodal framework for harmful meme detection.
The Hateful Memes detection challenge by Facebook \cite{kiela2020hateful} introduced the task of classifying a meme as hateful vs. non-hateful. 
Different approaches such as feature augmentation, attention mechanism, and multimodal loss re-weighting were attempted \cite{das2020detecting, sandulescu2020detecting, zhou2021multimodal, lippe2020multimodal} as part of this task. \citet{sabat2019hate} studied hateful memes by highlighting the importance of visual cues such as structural template, graphic modality, causal depiction, etc. 

Web-entity detection along with fair face classification \cite{karkkainen2019fairface} and semi-supervised learning-based classification \cite{zhong2020classification} were also used for the hateful meme classification task. 
Other noteworthy research includes using implicit models, e.g.,~topic modelling and multimodal cues, for detecting offensive analogy \cite{shang2021aomd} and hateful discrimination \cite{mittos2019and} in memes. \citet{wang2020understanding} argued that online attention can be garnered immensely via fauxtography, which could eventually evolve towards turning into memes that potentially go viral. 
To support research on these topics, several datasets for offensiveness, hate speech, and harmfulness detection have been created \cite{suryawanshi-etal-2020-multimodal, kiela2020hateful, pramanick-etal-2021-detecting, pramanick-etal-2021-momenta-multimodal, gomez2019exploring,ACL2021:propaganda:memes,Survey:2022:Harmful:Memes}.

Most of the above studies attempted to address classification tasks in a constrained setting. However, to the best of our knowledge, none of them targeted the task of detecting the specific entities that are being targeted. Here, we aim to bridge this gap with focus on detecting the specific entities targeted by a given harmful meme.

\begin{table}[t!]
\centering
\resizebox{0.8\columnwidth}{!}{%
\begin{tabular}{c|c|c|c}
\hline
\multirow{2}{*}{\textbf{Split}} & \multirow{2}{*}{\textbf{\# Examples}} & \multicolumn{2}{c}{\textbf{Category-wise \# Samples.}} \\ \cline{3-4} 
& & \textbf{Harmful} &\textbf{Not-harmful}\\ \hline
Train & 3,618 & 1,206 & 2,412 \\ 
Validation & 216 & 72 & 144 \\ 
Test & 612 & 316 & 296 \\ \hdashline
Total & 4,446 & 1,594 & 2,852 \\ \hline
\end{tabular}
}
\caption{Summary of \extharmp,\ with overall and category-wise \# of samples.}
\label{tab:datastat}
\end{table}
\section{Dataset}
The \olddata\ dataset \cite{pramanick-etal-2021-momenta-multimodal} consists of 3,552 memes about US politics. Each meme is annotated with its harmful label and the social entity that it targets. The targeted entities are coarsely classified into four social groups: individual, organization, community, and the general public. While these coarse classes provide an abstract view of the targets, identifying the \emph{specific} targeted person, organization, or community in a fine-grained fashion is also crucial, and this is our focus here. All the memes in this dataset broadly pertain to \textit{US Politics domain}, and they target well-known personalities or organizations. To this end, we manually re-annotated the memes in this dataset with the specific people, organizations, and communities that they target.         

\paragraph{Extending \olddata\ (\extharmp).}
Towards generalizability, we extend \olddata\ 
by redesigning the existing data splits as shown in Table~\ref{tab:datastat}. We call the resulting dataset \extharmp. It contains a total of 4,446 examples including 1,594 harmful and 2,852 non-harmful; both categories have references to a number of entities.
For training, we use the \textit{harmful} memes provided as part of the original dataset \cite{pramanick-etal-2021-momenta-multimodal}, which we re-annotate for the fine-grained entities that are being targeted harmfully as positive samples (\textit{harmful} targets). This is matched with twice as many negative samples (\textit{not-harmful} targets). For negative targets, we use the top-2 entities from the original entity lexicon, which are not labeled for harmfulness and have the highest lexical similarity with the meme text \cite{FERREIRA20161}. This at least ensures lexical similarity with the entities referenced within a meme, thereby facilitating a confounding effect \cite{kiela2020hateful} as well. For the test set, \textit{all} the entities are first extracted automatically using named entity recognition (NER) and person identification (PID)\footnote{NER using SpaCy \& PID using \url{http://github.com/ageitgey/face_recognition.}}. This is followed by manual annotation of the test set.

\begin{figure}[t!]
{\setlength{\fboxsep}{0pt}%
\framebox{%
\includegraphics[width=\columnwidth]{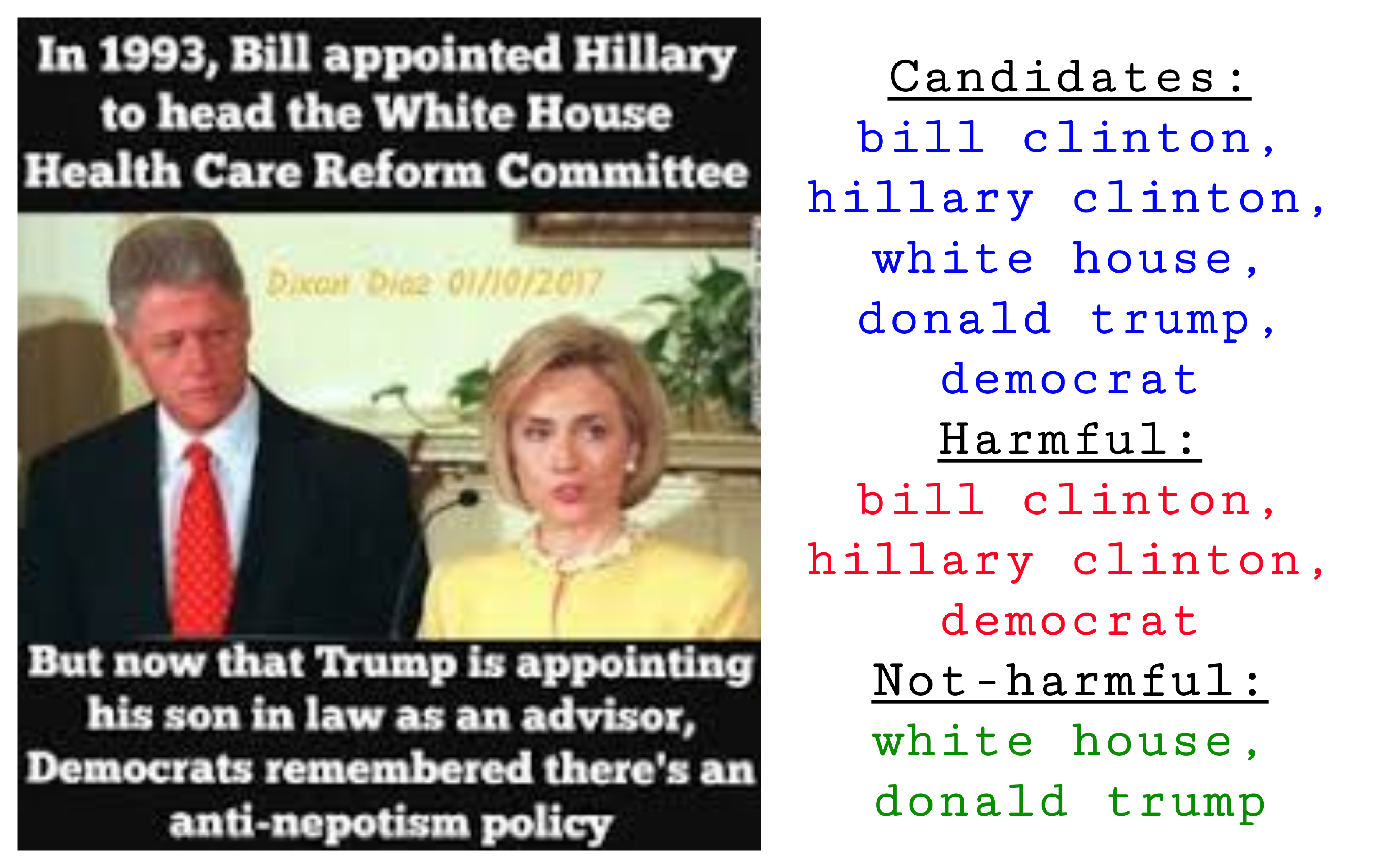}}}
\caption{Example meme, along with the \textcolor{blue}{candidate} entities, \textcolor{red}{harmful} targets, and \textcolor{darkgreen}{non-harmful} references.}
\label{fig:annot_ex}
\end{figure}

\begin{figure*}[ht!]
\centering
\subfloat[{\centering Individual}\label{fig:harm_ind}]{
\includegraphics[width=0.335\columnwidth,height=0.20\columnwidth]{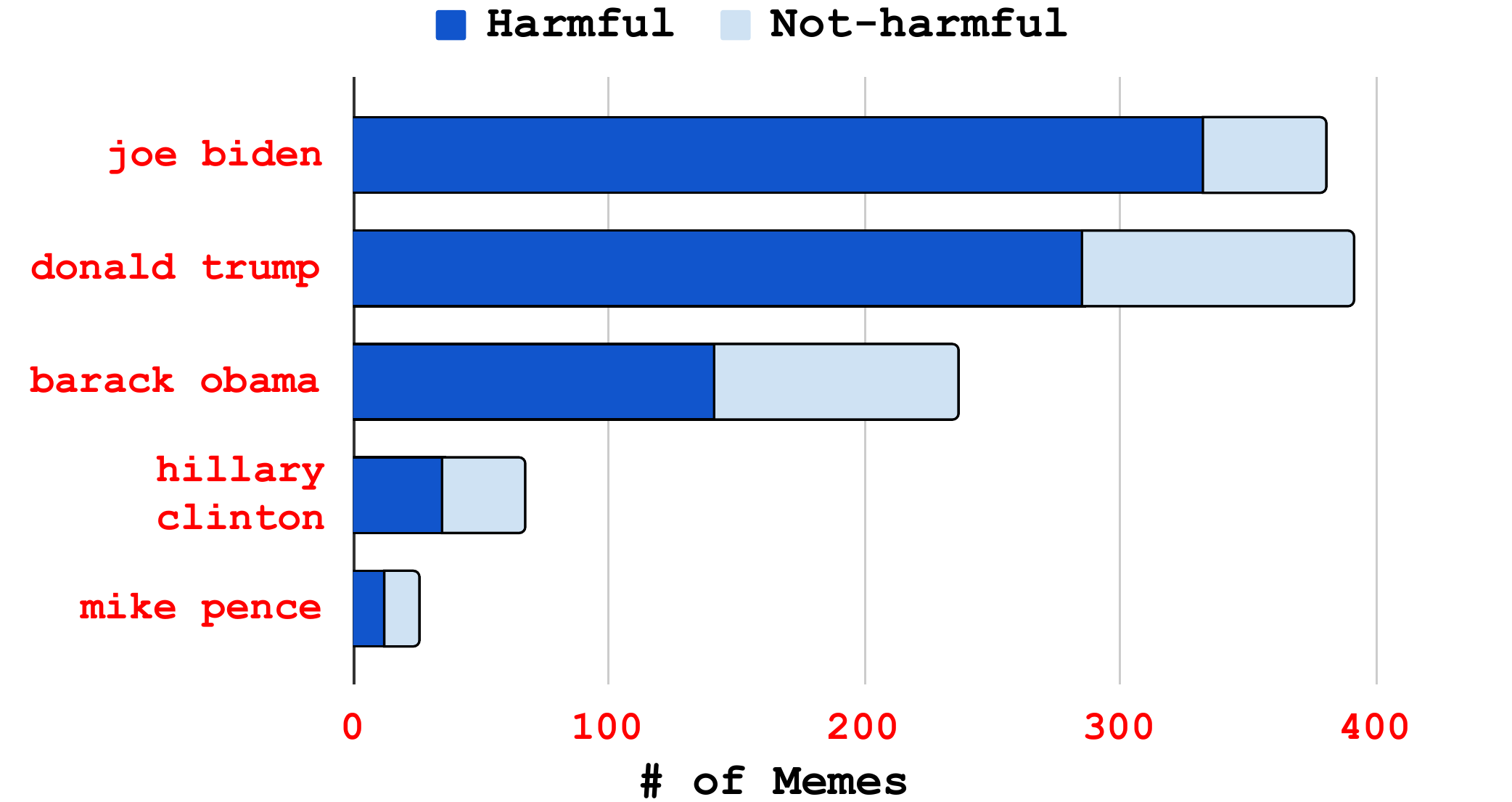}}
\subfloat[{Organization}\label{fig:harm_org}]{
\includegraphics[width=0.335\columnwidth,height=0.20\columnwidth]{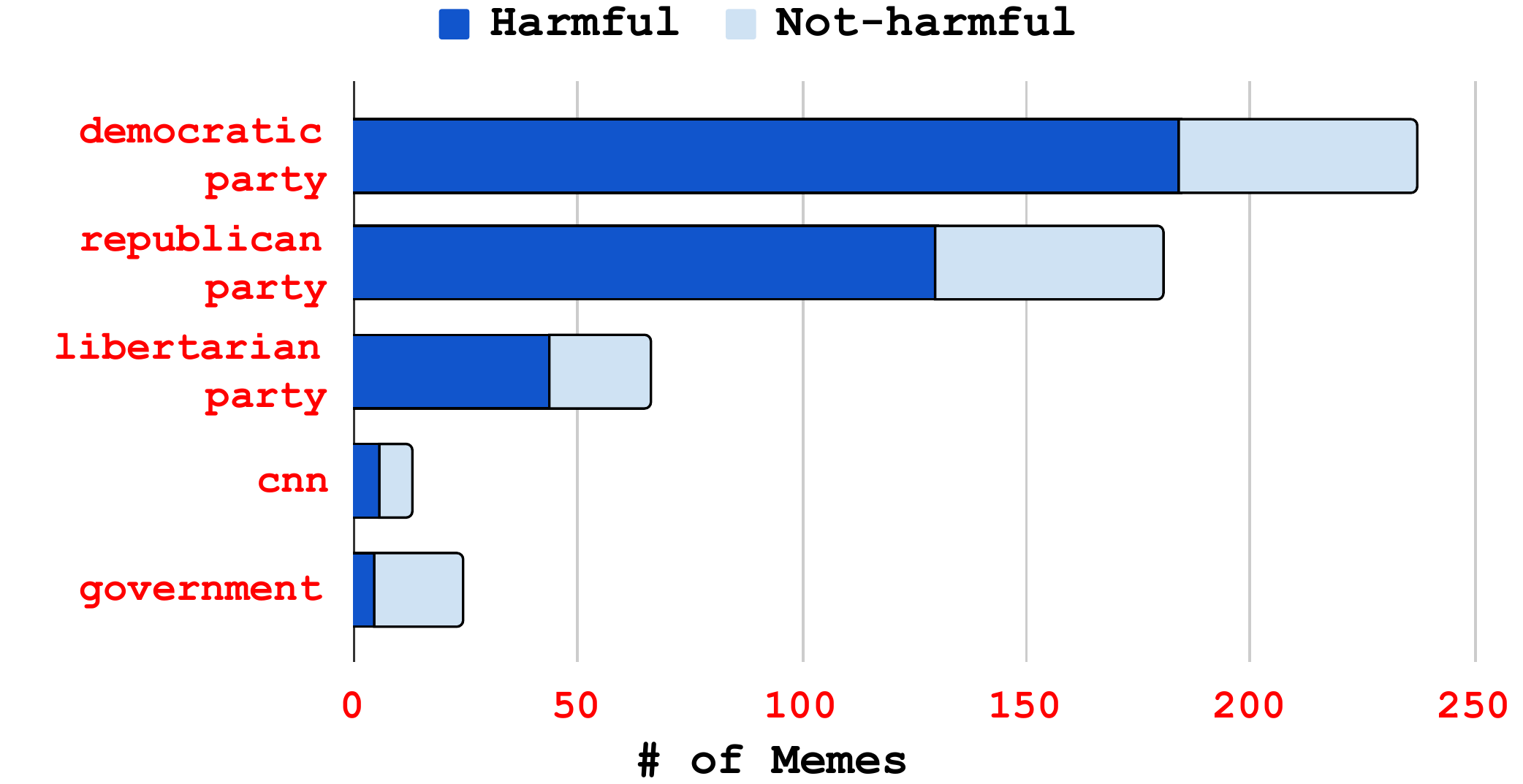}}
\subfloat[{\centering Community}\label{fig:harm_comm}]{
\includegraphics[width=0.335\columnwidth,height=0.20\columnwidth]{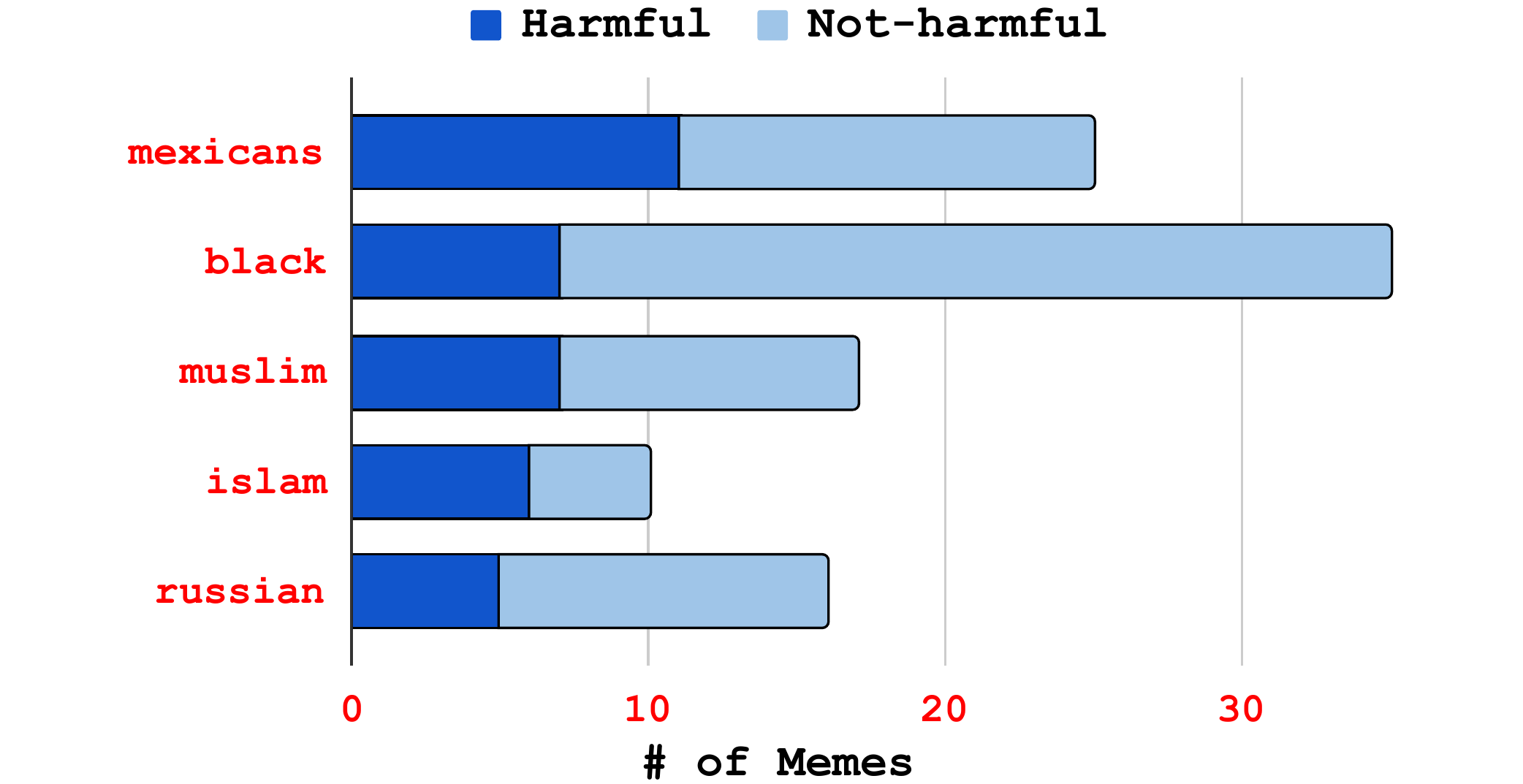}}
\caption{Comparison plots for the top-5 \textit{harmfully referenced} entities, for their harmful/non-harmful referencing in our dataset.}
\label{fig:top5harmful}
\end{figure*}

\paragraph{Dataset Annotation Process}
Since assessing the harmfulness of memes is a highly subjective task, our annotators were requested to follow four key steps when annotating each meme, aiming to ensure label consistency. The example in Fig.~\ref{fig:annot_ex} demonstrates the steps taken while annotating: we first identify the candidate entities, and then we decide whether a given entity is targeted in a harmful way. We asked our annotators to do the following (additional details about the annotation process are given in Appendix~\ref{sec:annotations}):

\begin{enumerate}
    \item Understand the meme and its background context.
    \item List all the valid \emph{candidate} entities that are referenced in the meme. For the example on Fig.~\ref{fig:annot_ex}, the valid entities are \emph{Bill Clinton, Hillary Clinton, White House, Donald Trump}, and \emph{Democrat}. 
    \item Assign the relevant entities as \emph{harmful}. For the example on Fig.~\ref{fig:annot_ex}, \emph{Bill Clinton, Hillary Clinton}, and \emph{Democrat} are targeted in the meme for influencing the appointment of their kin on government positions.  
    \item Finally, assign \emph{harmless} references to entities under the non-harmful category. In the example on Fig.~\ref{fig:annot_ex}, \emph{Donald Trump} and \emph{White House} would be annotated as \textit{non-harmful}.
\end{enumerate}

We had three annotators and a consolidator. The inter-annotator agreement before consolidation had a Fleiss Kappa of 0.48 (moderate agreement), and after consolidation it increased to 0.64 (substantial agreement).

\paragraph{Analyzing Harmful Targeting in Memes.}
The memes in \extharmp\ are about \emph{US Politics}, and thus they prominently feature entities such as \emph{Joe Biden} and \emph{Donald Trump}, both harmfully and harmlessly. The ratio between these types of referencing varies across \emph{individuals}, \emph{organizations}, and \emph{communities}. We can see in Fig.~\ref{fig:top5harmful} that the top-5 harmfully referenced \emph{individuals} and \emph{organizations} are observed to be subjected to a more relative harm (normalized by the number of occurrences of these entities in memes). 
However, the stacked plots for the top-5 harmfully targeted communities \emph{Mexicans, Black, Muslim, Islam}, and \emph{Russian} in Fig.~\ref{fig:harm_comm} show relatively less harm targeting these communities.


\begin{figure}[t!]
    \resizebox{\columnwidth}{!}{
    \centering
    \includegraphics{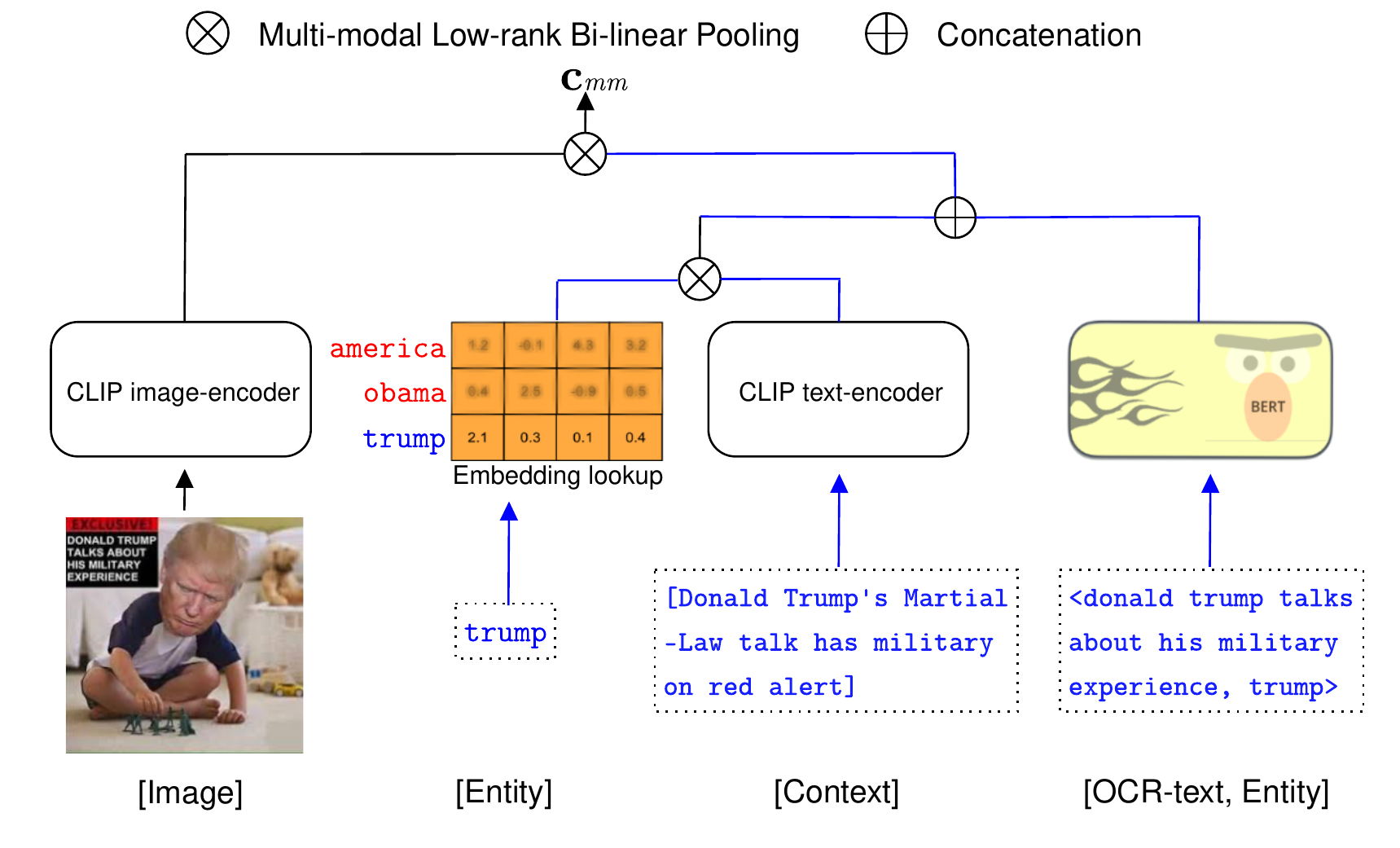}}
    \caption{The architecture of our proposed approach \model. Here, $\mathbf{c}_{mm}$ is the multimodal representation used for the final classification.}
    \label{fig:modelarch}
\end{figure}

\section{Proposed Approach}
Our proposed model \model,\ as depicted in Fig.~\ref{fig:modelarch}, is based on a fusion of the textual and the visual modalities, explicitly enriched via contextualised representations by leveraging CLIP \cite{radford2021learning}. We chose CLIP as a preferred encoder module for contextualization, due to its impressive zero-shot multimodal embedding capabilities. At first, valid entities are extracted automatically, as part of the process of creating training/validation sets. Then, for each meme, we first obtain the \emph{contextualized-entity} (\content) representation by fusing the CLIP-encoded context and the entity representation. \content\ is then fused with BERT-based \cite{devlin2019bert} \emph{embedded-harmfulness} (\embharm) encoding fine-tuned on the OCR-extracted text and entities as inputs. We call the resulting fusion output a \emph{contextualized-text} (\conttxt) representation. \conttxt\ is then fused with the \textit{contextualized-image} (\contimg) representation, obtained using the CLIP encoder for the image. We, henceforth, refer to the resulting enriched representation as the \textit{contextualized multimodal} (\contmm) representation. 
We modify the multimodal low-rank bi-linear pooling \cite{kim2017hadamard} to fuse the input representation into a  joint space.

This approach, as can be seen in the subsequent sections below, not only can capture complex cross-modal interactions, but it also provides an efficient fusion mechanism towards obtaining a context-enriched representation. Finally, we use this representation to train a classifier for our task. 
We describe each module in detail below.

\paragraph{Low-rank Bi-linear Pooling (LRBP).}
We begin by revisiting \textit{low-rank bi-linear pooling} to set the necessary background. Due to the many parameters in bi-linear models, \citet{NIPS2009147702db} suggested a low-rank bi-linear (LRB) approach to reduce the rank of the weight matrix $\mathbf{W}_i$. Consequently, the number of parameters and hence the complexity, are reduced. The weight matrix $\mathbf{W}_i$ is re-written as $\mathbf{W}_i = \mathbf{U}_i \mathbf{V}^{T}_i$, where $\mathbf{U}_i\in\mathbb{R}^{N\times d}$ and $\mathbf{V}_i\in\mathbb{R}^{M\times d}$, effectively putting an upper bound of $\min(N, M)$ on the value of $d$.
Therefore, the low-rank bi-linear models can be expressed as follows:
\begin{equation}
         \small f_i = \mathbf{x}^{T}\mathbf{W}_i\mathbf{y} = \mathbf{x}^{T}\mathbf{U}_i\mathbf{V}^{T}_i\mathbf{y}
         = \mathbbm{1}^{T}(\mathbf{U}^{T}_i\mathbf{x} \circ \mathbf{V}^{T}_i\mathbf{y})
    \label{eq:lrbm_order}
\end{equation}
where $\mathbbm{1}\in\mathbb{R}^{d}$ is a column vector of ones, and $\circ$ is Hadamard product. 
$f_i$ in Equation (\ref{eq:lrbm_order}) can be further re-written to obtain $\mathbf{f}$ as follows:
\begin{equation}
    \begin{array}{rcl}
    \mathbf{f} &=& \mathbf{P}^{T}(\mathbf{U}^{T}\mathbf{x} \circ \mathbf{V}^{T}\mathbf{y}) + \mathbf{b}
\end{array}
\label{eq:lrbm}
\end{equation}
where $\mathbf{f}\in\{f_i\}$, $\mathbf{P}\in\mathbb{R}^{d\times c}$, $\mathbf{b}\in\mathbb{R}^{c}$,
$d$ is an output, and $c$ is an LRB hyper-parameter.

We further introduce a non-linear activation formulation for LRBP, following \citet{kim2017hadamard}, who argued that non-linearity both before and after the Hadamard product complicates the gradient computation. This addition to Equation~(\ref{eq:lrbm}) can be represented as follows:
\begin{equation}
    \begin{array}{rcl}
    \mathbf{f} &=& \mathbf{P}^{T}\tanh(\mathbf{U}^{T}\mathbf{x} \circ \mathbf{V}^{T}\mathbf{y}) + \mathbf{b}
\end{array}
\label{eq:lrbp}
\end{equation}

We slightly modify the multimodal low-rank bi-linear pooling (\mmlrbp).\ Instead of directly projecting the input $\mathbf{x}\in\mathbb{R}^{N}$ and $\mathbf{y}\in\mathbb{R}^{M}$ into a lower dimension $d$, we first project the input modalities in a joint space $N$. We then perform LRBP as expressed in Equation \ref{eq:lrbp}, by using jointly embedded representations $\mathbf{x}_{mm}\in\mathbb{R}^{N\times d}$ and $\mathbf{y}_{mm}\in\mathbb{R}^{N\times d}$ to obtain a multimodal fused representation $\mathbf{f}_{mm}$, as expressed below:
\begin{equation}
    \begin{array}{rcl}
    \mathbf{f}_{mm} &=& \mathbf{P}^{T}\tanh(\mathbf{U}^{T}\mathbf{x}_{mm} \circ \mathbf{V}^{T}\mathbf{y}_{mm}) 
\end{array}
\label{eq:disarmlrbp}
\end{equation}
\paragraph{Structured Context.}
Towards modelling auxiliary knowledge, we curate \textit{contexts} for the memes in \extharmp.\ First, we use the meme text as a search query\footnote{\url{https://pypi.org/project/googlesearch-python/}}
to retrieve relevant contexts,
using the title and the first paragraph of the resulting top document as a \emph{context}, which we call $con$. 

\paragraph{Contextualized-entity Representation (CE).}
Towards modelling the context-enriched entity, we first obtain the embedding of the input entity $ent$. Since we have a finite set of entities referenced in the memes in our training dataset, we perform a lookup in the embedding matrix from $\mathbb{R}^{V\times H}$ to obtain the corresponding entity embedding $\mathbf{ent}\in\mathbb{R}^{H}$, with $H=300$ being the embedding dimension and $V$ the vocabulary size. We train the embedding matrix from \textit{scratch} as part of the overall training of our model.
We project the obtained entity representation $\mathbf{ent}$ into a 512-dimensional space, which we call $\mathbf{e}$. To augment a given entity with relevant contextual information, we fuse it with a contextual representation $\mathbf{c}\in\mathbb{R}^{512}$ obtained by encoding the associated context ($con$) using CLIP. We perform this fusion using our adaptation of the multimodal low-rank bi-linear pooling as defined by Equation~(\ref{eq:disarmlrbp}). This yields the following contextualized-entity (\content) representation $\mathbf{c}_{ent}$:
\begin{equation}
    \begin{array}{rcl}
    \mathbf{c}_{ent} &=& \mathbf{P}^{T}_{1}\tanh(\mathbf{U}^{T}_{1}\mathbf{e} \circ \mathbf{V}^{T}_{1}\mathbf{c}) + \mathbf{b}
\end{array}
\label{eq:mlrbpcent}
\end{equation}
where $\mathbf{c}_{ent}\in\mathbb{R}^{512}$, $\mathbf{P}_{1}\in\mathbb{R}^{256\times 512}$, $\mathbf{b}\in\mathbb{R}^{512}$, $\mathbf{U}_{1}\in\mathbb{R}^{512\times 256}$, and $\mathbf{V}_{1}\in\mathbb{R}^{512\times 256}$.

\paragraph{Contextualized-Text (CT) Representation.}
Once we obtain the contextualized-entity embedding $\mathbf{c}_{ent}$, we concatenate it with the BERT encoding for the combined representation of the OCR-extracted text and the entity ($\mathbf{o}_{ent}\in\mathbb{R}^{768}$). We call this encoding an \textit{embedded-harmfulness} (\embharm) representation. The concatenated representation from $\mathbb{R}^{1280}$ is then projected non-linearly into a lower dimension using a dense layer of size 512. 
We call the resulting vector $\mathbf{c}_{txt}$ a \emph{contextualized-text} (CT) representation:
\begin{equation}
    \begin{array}{ccc}
         \mathbf{c}_{txt}&=&\mathbf{W}_i[\mathbf{o}_{ent},\mathbf{c}_{ent}] + b_i
    \end{array}
\end{equation}
where $\mathbf{W}\in\mathbb{R}^{1280\times512 }$.

\paragraph{Contextualized Multimodal (CMM) Representation.}
Once we obtain the contextualized-text representation $\mathbf{c}_{txt}\in\mathbb{R}^{512}$, we again perform multimodal low-rank bi-linear pooling using Equation~(\ref{eq:disarmlrbp}) to fuse it with the contextualized-image representation $\mathbf{c}_{img}\in\mathbb{R}^{512}$, obtained using the CLIP image-encoder. The operation is expressed as follows:
\begin{equation}
    \begin{array}{rcl}
    \mathbf{c}_{mm} &=& \mathbf{P}^{T}_{2}\tanh(\mathbf{U}^{T}_{2}\mathbf{c}_{txt} \circ \mathbf{V}^{T}_{2}\mathbf{c}_{img}) 
\end{array}
\label{eq:mlrbpcmm}
\end{equation}
where $\mathbf{c}_{mm}\in\mathbb{R}^{512}$, $\mathbf{P}_{2}\in\mathbb{R}^{256\times 512}$, 
$\mathbf{U}_{2}\in\mathbb{R}^{512\times 256}$, and $\mathbf{V}_{2}\in\mathbb{R}^{512\times 256}$. 

Notably, we learn two different projection matrices $\mathbf{P}_{1}$ and $\mathbf{P}_{2}$, for the two fusion operations performed as part of Equations (\ref{eq:mlrbpcent}) and (\ref{eq:mlrbpcmm}), respectively, since the fused representations at the respective steps are obtained using different modality-specific interactions.

\paragraph{Classification Head.}
Towards modelling the binary classification for a given meme and a corresponding entity as either harmful or non-harmful, we use a shallow multi-layer perceptron with a single dense layer of size 256, which represents a condensed representation for classification. We finally map this layer to a single dimension output via a sigmoid activation. We use binary cross-entropy for the back-propagated loss.

\section{Experiments}
We experiment with various unimodal (image/text-only) and multimodal models, including such pre-trained on multimodal datasets such as MS COCO \cite{lin2014microsoft} and CC \cite{sharma2018conceptual}. We train \model~and all unimodal baselines using PyTorch, while for the multimodal baselines, we use the MMF framework.\footnote{\url{github.com/facebookresearch/mmf}} \footnote{Additional details along with the values of the hyper-parameters are given in Appendix \ref{sec:hyperparameters}.}

\subsection{Evaluation Measures}

For evaluation, we use commonly used macro-average versions of accuracy, precision, recall, and F1 score. For example, we discuss the harmful class recall, which is relevant for our study as it characterizes the model's performance at detecting \textit{harmfully} targeting memes. All results we report are averaged over five independent runs. 

\paragraph{Evaluation Strategy.}
With the aim of having a realistic setting, we pose our evaluation strategy as an open-class one. We train all systems using under-sampling of the entities that were not targeted in a harmful way: using all positive (harmful) examples and twice as many negative (non-harmful) ones. We then perform an open-class testing, for all referenced entities (some possibly unseen on training) per meme, effectively making the evaluation more realistic. To this end, we formulate three testing scenarios as follows, with their Harmful (H) and Non-harmful (N) counts: 
\begin{enumerate}
    \item \textbf{ Test set A (316H, 296N)}: All examples in this dataset are about entities that were \textit{seen} during training. 
    \item \textbf{ Test set B (27H, 94N)}: The examples in this set are about entities that were \textit{not seen} as \textit{harmful} during training. 
    \item \textbf{ Test set C (16H, 76N)}: All examples are about entities that were completely \textit{unseen} during training.
\end{enumerate}

\begin{table*}[htb!]
\centering
\resizebox{\textwidth}{!}{
\begin{tabular}{c|c|l|c|c|c|c|c|c|c|c||c|c|c|c|c|c|c|c}
\hline

\multirow{3}{*}{\bf System} & \multirow{3}{*}{\bf Modality} & \multirow{3}{*}{\bf Approach} & \multicolumn{8}{c||}{Test Set A} & \multicolumn{8}{c}{Test Set B} \\ \cline{4-19}
& & & \multirow{2}{*}{\textbf{Acc}} & \multirow{2}{*}{\textbf{Prec}} & \multirow{2}{*}{\textbf{Rec}} & \multirow{2}{*}{\textbf{F1}} & \multicolumn{2}{c|}{Not-harmful} & \multicolumn{2}{c||}{Harmful} & \multirow{2}{*}{\textbf{Acc}} & \multirow{2}{*}{\textbf{Prec}} & \multirow{2}{*}{\textbf{Rec}} & \multirow{2}{*}{\textbf{F1}} & \multicolumn{2}{c|}{Not-harmful} & \multicolumn{2}{c}{Harmful} \\ \cline{8-11} \cline{16-19}

& & & & & & & \textbf{P} & \textbf{R} & \textbf{P} & \textbf{R} & & & & & \textbf{P} & \textbf{R} & \textbf{P} & \textbf{R} \\ \hline

 & & XLNet Text-only & 0.6765 & 0.69 & 0.67 & 0.6663 & \multicolumn{1}{l|}{0.73} & 0.52 & \multicolumn{1}{l|}{0.65} & 0.82 & 0.5041 & 0.425 & 0.405 & 0.4060 & \multicolumn{1}{c|}{0.72} & 0.59 & \multicolumn{1}{c|}{0.13} & 0.22 \\ 
 & & VGG Image-only & 0.7451 & 0.75 & 0.745 & 0.7438 & \multicolumn{1}{l|}{0.71} & 0.81 & \multicolumn{1}{l|}{0.79} & 0.68 & 0.5455 & 0.42 & 0.405 & 0.4101 & \multicolumn{1}{c|}{0.73} & 0.66 & \multicolumn{1}{c|}{0.11} & 0.15\\ 
 &  & GRU Text-only & 0.7484 & 0.745 & 0.75 & 0.7473 & \multicolumn{1}{l|}{0.73} & 0.76 & \multicolumn{1}{l|}{0.76} & 0.74 & 0.5455 & 0.43 & 0.42 & 0.4210 & \multicolumn{1}{c|}{0.73} & 0.65 & \multicolumn{1}{c|}{0.13} & 0.19\\ 
& \multirow{-4}{*}{\rotatebox{90}{Unimodal}} & VIT Image only & 0.7647 & 0.765 & 0.765 & 0.7642 & \multicolumn{1}{l|}{0.74} & 0.79 & \multicolumn{1}{l|}{0.79} & 0.74 & 0.5207 & 0.525 & 0.535 & 0.4843 & \multicolumn{1}{c|}{0.8} & 0.51 & \multicolumn{1}{c|}{0.25} & 0.56\\ \cline{2-19}

 & & ViLBERT CC & 0.6895 & 0.69 & 0.685 & 0.6835 & \multicolumn{1}{l|}{0.71} & 0.6 & \multicolumn{1}{l|}{0.67} & 0.77& 0.438 & 0.535 & 0.53 & 0.4302 & \multicolumn{1}{c|}{0.82} & 0.35 & \multicolumn{1}{c|}{0.25} & \textbf{0.71} \\ 
 & & MM Transformer & 0.6993 & 0.71 & 0.695 & 0.6926 & \multicolumn{1}{l|}{0.75} & 0.57 & \multicolumn{1}{l|}{0.67} & 0.82 & \textbf{0.7769} & 0.53 & 0.575 & 0.5032 & \multicolumn{1}{c|}{0.78} & 0.51 & \multicolumn{1}{c|}{0.28} & 0.64 \\ 
 & & VisualBERT & 0.7026 & 0.725 & 0.69 & 0.6918 & \multicolumn{1}{l|}{\textbf{0.78}} & 0.54 & \multicolumn{1}{l|}{0.67} & 0.84 & 0.5537 & 0.545 & 0.565 & 0.5108 & \multicolumn{1}{c|}{0.82} & 0.54 & \multicolumn{1}{c|}{0.27} & 0.59 \\ 
 & & VisualBERT – COCO & 0.7059 & 0.71 & 0.7 & 0.7014 & \multicolumn{1}{l|}{0.73} & 0.62 & \multicolumn{1}{l|}{0.69} & 0.78 & 0.5785 & 0.53 & 0.545 & 0.5147 & \multicolumn{1}{c|}{0.8} & 0.61 & \multicolumn{1}{c|}{0.26} & 0.48 \\ 
 & & MMBT & 0.7157 & 0.72 & 0.71 & 0.7121 & \multicolumn{1}{l|}{0.74} & 0.64 & \multicolumn{1}{l|}{0.7} & 0.78 & 0.6116 & 0.54 & 0.55 & 0.5310 & \multicolumn{1}{c|}{0.81} & 0.66 & \multicolumn{1}{c|}{0.27} & 0.44 \\ 
\multirow{-9}{*}{\rotatebox{90}{Baselines}} & & ViLBERT & 0.7516 & 0.755 & 0.75 & 0.7495 & \multicolumn{1}{l|}{\textbf{0.78}} & 0.68 & \multicolumn{1}{l|}{0.73} & 0.82 & 0.6612 & 0.58 & 0.595 & 0.5782 & \multicolumn{1}{c|}{\textbf{0.83}} & 0.71 & \multicolumn{1}{c|}{0.33} & 0.48 \\ \cline{1-1} \cline{3-19}

& & \content\ + \contimg\ (concat) & 0.7353 & 0.74 & 0.735 & 0.7361 & \multicolumn{1}{l|}{0.71} & 0.77 & \multicolumn{1}{l|}{0.77} & 0.7 & 0.4793 & 0.46 & 0.44 & 0.4230 & \multicolumn{1}{c|}{0.74} & 0.51 & \multicolumn{1}{c|}{0.18} & 0.37 \\ 
& & \content\ + \contimg\ (\mmlrbp) & \textbf{0.781} & \textbf{0.785} & 0.78 & 0.7790 & \multicolumn{1}{l|}{0.74} & \textbf{0.84} & \multicolumn{1}{l|}{\textbf{0.83}} & 0.72 & 0.562 & 0.535 & 0.545 & 0.5079 & \multicolumn{1}{c|}{0.81} & 0.57 & \multicolumn{1}{c|}{0.26} & 0.52 \\ 
& & \embharm\ + \contimg\ (concat) & 0.6634 & 0.665 & 0.66 & 0.6609 & \multicolumn{1}{l|}{0.67} & 0.6 & \multicolumn{1}{l|}{0.66} & 0.72 & 0.5868 & 0.505 & 0.51 & 0.4964 & \multicolumn{1}{c|}{0.78} & 0.65 & \multicolumn{1}{c|}{0.23} & 0.37 \\ 
& & \embharm\ + \contimg\ (\mmlrbp) & 0.7255 & 0.73 & 0.725 & 0.7260 & \multicolumn{1}{l|}{0.74} & 0.67 & \multicolumn{1}{l|}{0.72} & 0.78 & 0.6612 & 0.545 & 0.555 & 0.5470 & \multicolumn{1}{c|}{0.8} & 0.74 & \multicolumn{1}{c|}{0.29} & 0.37 \\ 
\multirow{-5}{*}{\rotatebox{90}{\begin{tabular}[c]{@{}c@{}}Prop. system \\ \& variants\end{tabular}}} & \multirow{-11}{*}{\rotatebox{90}{Multimodal}} & {\color[HTML]{000000} \model} & {\color[HTML]{000000} \textbf{0.781}} & {\color[HTML]{000000} 0.74} & {\color[HTML]{000000} \textbf{0.835}} & {\color[HTML]{000000} \textbf{0.7845}} & \multicolumn{1}{l|}{{\color[HTML]{000000} 0.74}} & {\color[HTML]{000000} 0.81} & \multicolumn{1}{l|}{{\color[HTML]{000000} 0.74}} & {\color[HTML]{000000} \textbf{0.86}} & {\color[HTML]{000000} 0.74} & {\color[HTML]{000000} \textbf{0.605}} & {\color[HTML]{000000} \textbf{0.74}} & {\color[HTML]{000000} \textbf{0.6498}} & \multicolumn{1}{c|}{{\color[HTML]{000000} \textbf{0.83}}} & {\color[HTML]{000000} \textbf{0.79}} & \multicolumn{1}{c|}{{\color[HTML]{000000} \textbf{0.38}}} & {\color[HTML]{000000} 0.69} \\ \hline
\multicolumn{3}{c|}{$\Delta_{(\model\ - ViLBERT)\times 100}(\%)$} & \textcolor{blue}{$\uparrow2.94\%$} & \textcolor{red}{$\downarrow1.5\%$} & \textcolor{blue}{$\uparrow8\%$} & \textcolor{blue}{$\uparrow3.5\%$} & \multicolumn{1}{c|}{\textcolor{red}{$\downarrow4\%$}} & \textcolor{blue}{$\uparrow13\%$} & \textcolor{blue}{$\uparrow1\%$} & \textcolor{blue}{$\uparrow4\%$} & \textcolor{blue}{$\uparrow7.88\%$} & \textcolor{blue}{$\uparrow2.5\%$} & \textcolor{blue}{$\uparrow14.5\%$} & \textcolor{blue}{$\uparrow7.16\%$} & \multicolumn{1}{c|}{--} & \textcolor{blue}{$\uparrow8\%$} & \textcolor{blue}{$\uparrow5\%$} & \multicolumn{1}{|c}{\textcolor{blue}{$\uparrow21\%$}} \\\hline
\end{tabular}}
\caption{Performance comparison of unimodal and multimodal models vs. \model\ (and its variants) on Test Sets A and B.}
\label{tab:ARes}
\end{table*}

\paragraph{Baseline Models.} Our baselines include both unimodal and multimodal models as follows: 
\begin{itemize}[leftmargin=*]
    \item[--] \underline{\em Unimodal Systems}: $\blacktriangleright$ {\bf VGG16, VIT:} For the unimodal (image-only) systems, we use two well-known models: VGG16 \cite{simonyan2015deep} and VIT (Vision Transformers) that emulate a Transformer-based application jointly over textual tokens and image patches \cite{dosovitskiy2021image}.
    $\blacktriangleright$ \textbf{GRU, XLNet:} For the unimodal (text-only) systems, we use GRU \cite{cho-etal-2014-learning}, which adaptively captures temporal dependencies, and XLNet \cite{yang2020xlnet}, which implements a generalized auto-regressive pre-training strategy.
    \item[--] \underline{\em Multimodal Systems}: $\blacktriangleright$ \textbf{MMF Transformer:} This is a multimodal Transformer model that uses visual and language tokens with self-attention.\footnote{\label{fn:mmf}\url{http://mmf.sh/docs/notes/model_zoo}}
    $\blacktriangleright$ \textbf{MMBT:} Multimodal Bitransformer \cite{kiela2020supervised} captures the intra-modal and the inter-modal dynamics.
    $\blacktriangleright$ \textbf{ViLBERT CC:} Vision and Language BERT \cite{lu2019vilbert}, pre-trained on CC \cite{sharma2018conceptual}, is a strong model with task-agnostic joint representation.  
    $\blacktriangleright$ \textbf{Visual BERT COCO:} Visual BERT \citep{li2019visualbert}, pre-trained on the MS COCO dataset \citep{lin2014microsoft}.
\end{itemize}

\begin{table}[htb!]
\centering
\resizebox{\columnwidth}{!}{
\begin{tabular}{c|c|l|c|c|c|c|cc|cc}
\hline
 & & \multicolumn{1}{c|}{} & &  &  &  & \multicolumn{2}{c|}{\textbf{Not-harmful}} & \multicolumn{2}{c}{\textbf{Harmful}} \\ \cline{8-11} 
\multirow{-2}{*}{\textbf{Sys}} & \multirow{-2}{*}{\textbf{}} & \multicolumn{1}{c|}{\multirow{-2}{*}{\textbf{Approach}}} & \multirow{-2}{*}{\textbf{Acc}} & \multirow{-2}{*}{\textbf{Prec}} & \multirow{-2}{*}{\textbf{Rec}} & \multirow{-2}{*}{\textbf{F1}} & \multicolumn{1}{c|}{\textbf{P}} & \textbf{R} & \multicolumn{1}{c|}{\textbf{P}} & \textbf{R} \\ \hline
 & & GRU Text-only & 0.478 & 0.45 & 0.41 & 0.394 & \multicolumn{1}{c|}{0.78} & 0.51 & \multicolumn{1}{c|}{0.12} & 0.31 \\ 
 & & VIT Image only & 0.532 & 0.435 & 0.4 & 0.403 & \multicolumn{1}{c|}{0.78} & 0.61 & \multicolumn{1}{c|}{0.09} & 0.19 \\ 
 & & XLNet Text-only & 0.445 & 0.51 & 0.515 & 0.415 & \multicolumn{1}{c|}{0.84} & 0.41 & \multicolumn{1}{c|}{0.18} & 0.62 \\ 
& \multirow{-4}{*}{\rotatebox{90}{Unimodal}} & VGG Image-only & 0.532 & 0.45 & 0.42 & 0.414 & \multicolumn{1}{c|}{0.79} & 0.59 & \multicolumn{1}{c|}{0.11} & 0.25 \\ \cline{2-11}
 & & ViLBERT CC & 0.358 & 0.53 & 0.49 & 0.350 & \multicolumn{1}{c|}{0.87} & 0.26 & \multicolumn{1}{c|}{0.19} & \textbf{0.72} \\ 
 & & VisualBERT & 0.478 & 0.535 & 0.56 & 0.442 & \multicolumn{1}{c|}{0.87} & 0.43 & \multicolumn{1}{c|}{0.2} & 0.69 \\ 
 & & MM Transformer & 0.510 & 0.505 & 0.505 & 0.448 & \multicolumn{1}{c|}{0.83} & 0.51 & \multicolumn{1}{c|}{0.18} & 0.5 \\ 
 & & ViLBERT & 0.608 & 0.525 & 0.54 & 0.505 & \multicolumn{1}{c|}{0.84} & 0.64 & \multicolumn{1}{c|}{0.21} & 0.44 \\ 
 & & VisualBERT – COCO & \textbf{0.771} & 0.525 & 0.515 & 0.511 & \multicolumn{1}{c|}{0.83} & \textbf{0.91} & \multicolumn{1}{c|}{0.22} & 0.12 \\ 
\multirow{-10}{*}{\rotatebox{90}{Baselines}} & & MMBT & 0.587 & 0.55 & 0.575 & 0.514 & \multicolumn{1}{c|}{0.87} & 0.59 & \multicolumn{1}{c|}{0.23} & 0.56 \\ \cline{1-1} \cline{3-11}
 & & \content\ + \contimg\ (concat) & 0.456 & 0.495 & 0.495 & 0.412 & \multicolumn{1}{c|}{0.82} & 0.43 & \multicolumn{1}{c|}{0.17} & 0.56 \\ 
 & & \content\ + \contimg\ (\mmlrbp) & 0.532 & 0.55 & 0.595 & 0.485 & \multicolumn{1}{c|}{\textbf{0.88}} & 0.5 & \multicolumn{1}{c|}{0.22} & 0.69 \\ 
 & & \embharm\ + \contimg\ (concat) & 0.532 & 0.48 & 0.475 & 0.442 & \multicolumn{1}{c|}{0.81} & 0.57 & \multicolumn{1}{c|}{0.15} & 0.38 \\ 
 & & \embharm\ + \contimg\ (\mmlrbp) & 0.619 & 0.5 & 0.495 & 0.483 & \multicolumn{1}{c|}{0.83} & 0.68 & \multicolumn{1}{c|}{0.17} & 0.31 \\ 
 \multirow{-5}{*}{\rotatebox{90}{\begin{tabular}[c]{@{}c@{}}Prop. system \\ \& variants\end{tabular}}} & \multirow{-10}{*}{\rotatebox{90}{Multimodal}} & {\color[HTML]{000000} \model} & {\color[HTML]{000000} 0.739} & {\color[HTML]{000000} \textbf{0.61}} & {\color[HTML]{000000} \textbf{0.73}} & {\color[HTML]{000000} \textbf{0.641}} & \multicolumn{1}{c|}{{\color[HTML]{000000} 0.86}} & {\color[HTML]{000000} 0.76} & \multicolumn{1}{c|}{{\color[HTML]{000000} \textbf{0.36}}} & {\color[HTML]{000000} 0.7} \\ \hline
 \multicolumn{3}{c|}{$\Delta_{(\model\ - MMBT)\times 100}(\%)$} & \textcolor{blue}{$\uparrow15.21\%$}
& \textcolor{blue}{$\uparrow6\%$}
& \textcolor{blue}{$\uparrow15.5\%$}
& \textcolor{blue}{$\uparrow12.66\%$}
& \multicolumn{1}{c|}{\textcolor{red}{$\downarrow1\%$}}
& \textcolor{blue}{$\uparrow17\%$}
& \textcolor{blue}{$\uparrow13\%$}
& \multicolumn{1}{|c}{\textcolor{blue}{$14\%$}}\\\hline
\end{tabular}}
\caption{Performance comparison of unimodal and multimodal models vs. \model\ (and its variants) on Test Set C.}
\label{tab:CRes}
\end{table}
\paragraph{Experimental Results.}
We compare the performance of several unimodal and multimodal systems (pre-trained or trained from scratch) vs. \model\ and its variants. All systems are evaluated using the 3-way testing strategy described above. We then perform ablation studies on representations that use the \textit{contextualized-entity}, its fusion with \textit{embedded-harmfulness} resulting into \textit{contextualized-text}, and the final fusion with \textit{contextualized-image} yielding the \textit{contextualized-multimodal} modules of \model\ 
(see Appendix~\ref{sec:ablation} for a detailed ablation study).\footnote{We use the abbreviations \content,\ \conttxt,\ \contimg,\ \contmm,\ \embharm,\ and \mmlrbp\ for the \textit{contextualized} representations of the entity, the text, the image, the multimodal representation, the embedded-harmfulness, and the multimodal low-rank bi-linear pooling, respectively.} This is followed by interpretability analysis. Finally, we discuss the limitations of \model\ by performing error analysis (details in Appendix~\ref{sec:err_analysis}).

\noindent {\bf \em All Entities Seen During Training}:
In our unimodal text-only baseline experiments, 
the GRU-based system yields a relatively lower \textit{harmful} recall of 0.74 compared to XLNet's 0.82, but a better overall F1 score of 0.75 vs. 0.67 for XLNet, as shown in Table \ref{tab:ARes}. The lower \textit{harmful} precision of 0.65 and the \textit{not-harmful} recall of 0.52 contribute to the lower F1 score for XLNet. 

Among the image-only unimodal systems, VGG performs better with a \textit{non-harmful} recall of 0.81, but its poor performance for detecting harmful memes yields a lower \textit{harmful} recall of 0.68. At the same time, VIT has a relatively better \textit{harmful} recall of 0.74. Overall, the unimodal results (see Table~\ref{tab:ARes}) indicate the efficacy of self-attention over convolution for images and RNN (GRU) sequence modeling for text.

Multimodally pre-trained models such as VisualBERT and ViLBERT yield moderate F1 scores of 0.70 and 0.68, and \textit{harmful} recall of 0.78 and 0.77, respectively (see Table \ref{tab:ARes}). Fresh training facilitates more meaningful results in favour of \textit{non-harmful} precision of 0.78 for both models, and \textit{harmful} recall of 0.84 and 0.82 for VisualBERT and ViLBERT, respectively. Overall, ViLBERT yields the most balanced performance of 0.75 in terms of F1 score. It can be inferred from these results (see Table \ref{tab:ARes}) that multimodal pre-training leverages domain relevance.

We can see in Table \ref{tab:ARes} that multimodal low-rank bi-linear pooling distinctly enhances the performance in terms of F1 score. The improvements can be attributed to the fusion of the \content\ and \embharm\ representations, respectively, with \contimg,\ instead of a simple concatenation. This is more prominent for \content\ with an F1 score of 0.78, which shows the importance of modeling the background context. Finally, \model\ yields a balanced F1 score of 0.78, with a reasonable precision of 0.74 for \textit{non-harmful} category, and the best recall of 0.86 for the \textit{harmful} category. 


\begin{figure*}[htb!]
\centering
\resizebox{\textwidth}{!}{%
\subfloat[{L-AT}\label{fig:comp_text}]{
\includegraphics[height=0.25\textwidth]{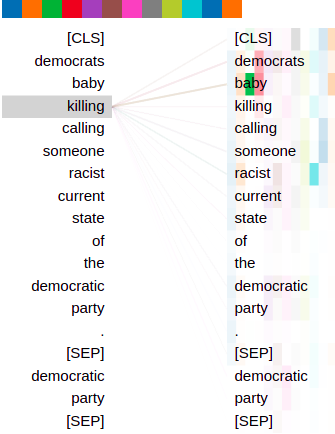}}\hspace{0.1mm}
\subfloat[{MM-AT-CLIP}\label{fig:comp_clip}]{
\includegraphics[width=0.25\textwidth, height=0.25\textwidth]{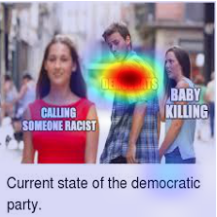}}\hspace{0.1mm}
\subfloat[{V-AT-\model}\label{fig:comp_disarm}]{
\includegraphics[width=0.25\textwidth, height=0.25\textwidth]{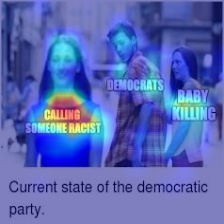}}\hspace{0.1mm}
\subfloat[{V-AT-ViLBERT}\label{fig:comp_vilbert}]{
\includegraphics[width=0.25\textwidth, height=0.25\textwidth]{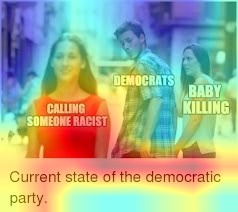}}}\\\vspace{0.4mm} 
\resizebox{\textwidth}{!}{%
\subfloat{\framebox[\textwidth]{\parbox{\dimexpr\linewidth-2\fboxsep-2\fboxrule}{Target Candidate$\rightarrow$\textcolor{blue}{\texttt{democratic party}} \justify{Context$\rightarrow$\textcolor{blue}{{Politics tears families apart during bruising political season, when many Americans drop friends and family members who have different political views.}}}}}}}
\caption{Comparison of the attention-maps for \model\ [(a), (b) \& (c)] and ViLBERT [(d)] using BertViz and Grad-CAM.}
\label{fig:comp_att}
\end{figure*}

\noindent  {\bf \em All Entities Unseen as Harmful Targets During Training}: 
With Test Set B, the evaluation is slightly more challenging in terms of the entities to be assessed, as these were never seen at training time as \emph{harmful}.

Unimodal systems perform poorly on the \textit{harmful} class, with the exception of XLNet (see Table~\ref{tab:ARes}), where the \textit{harmful} class recall as 0.56. For the multimodal baselines, systems pre-trained using COCO (VisualBERT) and CC (ViLBERT) yield a moderate recall of 0.64 and 0.71 for the \textit{harmful} class in contrast to what we saw for Test Set A in Table~\ref{tab:ARes}. This could be due to additional common-sense reasoning helping such systems, on a test set that is more open-ended compared to Test Set A. Their non-pre-trained versions along with the MM Transformer and MMBT achieve better F1 scores, but with low \textit{harmful} recall. 

Multimodal fusion using \mmlrbp\ improves the \textit{harmful} class recall for \content\ to 0.52, but yields lower values of 0.37 for \embharm\ fusion with \contimg\ (see Table \ref{tab:ARes}). This reconfirms the utility of the context. In comparison, \model\ yields a balanced F1 score of 0.65 with the best precision of 0.83 and 0.38, along with decent recall of 0.79 and 0.69 for \textit{non-harmful} and \textit{harmful} memes, respectively.

\noindent {\bf \em All Entities Unseen During Training}:
The results decline in this scenario (similarly to Test Set B), except for the \textit{harmful} class recall of 0.62 for XLNet, as shown in Table~\ref{tab:CRes}. In the current scenario (Test Set C), none of the entities being assessed at testing is seen during the training phase. 
For multimodal baselines, we see a similar trend for VisualBERT (COCO) and ViLBERT (CC), with the \textit{harmful} class recall of 0.72 for ViLBERT (CC) being significantly better than the 0.12 for VisualBERT (COCO). This again emphasizes the need for the affinity between the pre-training dataset and the downstream task at hand. In general, the precision for the \textit{harmful} class is very low.

We observe (see Table \ref{tab:CRes}) sizable boost for the \textit{harmful} class recall for \mmlrbp-based multimodal fusion of \contimg\ with \content\ (0.69\%), against a decrease with \embharm\ (0.31\%). In comparison, \model\ yields a low, yet the best \textit{harmful} precision of 0.36, and a moderate recall of 0.70 (see Table~\ref{tab:CRes}). Moreover, besides yielding reasonable precision and recall of 0.86 and 0.76 for the \textit{non-harmful} class, \model\ achieves better average precision, recall, and F1 scores of 0.61, 0.73, and 0.64, respectively.
\paragraph{Generalizability of DISARM.}
The generalizability of \model\ follows from its characteristic modelling and context-based fusion. \model\ demonstrates an ability to detect harmful targeting for a diverse set of entities. Specifically, the three-way testing setup inherently captures the efficacy with which \model\ can detect \textit{unseen} harmful targets. The prediction for entities \textit{completely} unseen on training yields better results (see Tables~\ref{tab:ARes} and \ref{tab:CRes}), and suggests possibly induced bias in the former scenario. 
Moreover, it is a direct consequence of the fact that we were able to incorporate only a limited set of the 246 potential targets. 
Overall, we argue that \model\ generalizes well for unseen entities with 0.65 and 0.64 macro-F1 scores, as compared to ViLBERT's 0.58 and MMBT's 0.51, for Test Sets B and C, respectively.

\paragraph{Comparative Diagnosis.}

Despite the marginally better \textit{harmful} recall for ViLBERT (CC) on Test Set B (see Table~\ref{tab:ARes}) and Test Set C (see Table~\ref{tab:CRes}), the overall balanced performance of \model\ appears to be reasonably justified based on the comparative interpretability analysis between the attention maps for the two systems.

Fig. \ref{fig:comp_att} shows the attention maps for an example meme. It depicts a meme that is \textit{correctly} predicted to \textit{harmfully} target the \textit{Democratic Party} by \model\ and incorrectly by ViLBERT. As visualised in Fig.~\ref{fig:comp_text}, the harmfully-inclined word \textit{killing} effectively attends not only to \textit{baby}, but also to \textit{Democrats} and \textit{racist}. The relevance is depicted via different color schemes and intensities, respectively. Interestingly, \textit{killing} also attends to the \textit{Democratic Party}, both as part of the OCR-extracted text and the target-candidate, jointly encoded by BERT. 
The multimodal attention leveraged by \model\ is depicted (via the CLIP encoder) in Fig.~\ref{fig:comp_clip}, demonstrating the utility of contextualised attention over the \textit{male} figure that represents an attack on the \textit{Democratic Party}. Also, \model\ has a relatively focused field of vision, as shown in Fig.~\ref{fig:comp_disarm}, as compared to a relatively scattered one for ViLBERT (see Fig.~\ref{fig:comp_vilbert}). This suggest a better multimodal modelling capacity for \model\ as compared to ViLBERT.

\section{Conclusion and Future Work}

We introduced the novel task of detecting the targeted entities within harmful memes and we highlighted the inherent challenges involved. Towards addressing this open-ended task, we extended \olddata\ with target entities for each harmful meme. We then proposed a novel multimodal deep neural framework, called \model, which uses an adaptation of multimodal low-rank bi-linear pooling-based fusion strategy at different levels of representation abstraction. 
We showed that \model\ outperforms various uni/multi-modal baselines in three different scenarios by 4\%, 7\%, and 13\% increments in terms of macro-F1 score, respectively. Moreover, \model\ achieved a relative error rate reduction of 9\% over the best baseline. We further emphasized the utility of different components of \model\ through ablation studies. We also elaborated on the generalizability of \model, thus confirming its modelling superiority over ViLBERT via interpretability analysis. We finally analysed the shortcomings in \model\ that lead to incorrect harmful target predictions.

In the present work, we made an attempt to elicit some inherent challenges pertaining to the task at hand: augmenting the relevant context, effectively fusing multiple modalities, and pre-training. Yet, we also leave a lot of space for future research for this novel task formulation.

\section*{Ethics and Broader Impact}

\paragraph{Reproducibility.}  We present detailed hyper-parameter configurations in Appendix~\ref{sec:hyperparameters} and Table~\ref{tab:hyperparameters}. 
The source code, and the dataset \extharmp\ are available at \url{https://github.com/LCS2-IIITD/DISARM}
 
\paragraph{User Privacy.} 
The information depicted/used does not include any personal information. Copyright aspects are attributed to the dataset source.  


\paragraph{Annotation.}

The annotation was conducted by NLP experts or linguists in India, who were fairly treated and were duly compensated. We conducted several discussion sessions to make sure all annotators could understand the distinction between harmful vs. non-harmful referencing.

\paragraph{Biases.}
Any biases found in the dataset are unintentional, and we do not intend to cause harm to any group or individual. We acknowledge that detecting harmfulness can be subjective, and thus it is inevitable that there would be biases in our gold-labelled data or in the label distribution. This is addressed by working on a dataset that is created using general keywords about US Politics, and also by following a well-defined schema, which sets explicit definitions for annotation. 

\paragraph{Misuse Potential.}
Our dataset can be potentially used for ill-intended purposes, such as biased targeting of individuals/communities/organizations, etc. that may or may not be related to demographics and other information within the text. Intervention with human moderation would be required to ensure that this does not occur.


\paragraph{Intended Use.}

We make use of the existing dataset in our work in line with the intended usage prescribed by its creators and solely for research purposes. This applies in its entirety to its further usage as well. We commit to releasing our dataset aiming to encourage research in studying harmful targeting in memes on the web. We distribute the dataset for research purposes only, without a license for commercial use. We believe that it represents a useful resource when used appropriately.

\paragraph{Environmental Impact.}

Finally, large-scale models require a lot of computations, which contribute to global warming \cite{strubell2019energy}. However, in our case, we do not train such models from scratch; rather, we fine-tune them on a relatively small dataset.

\section*{Acknowledgments}
The work was partially supported by a Wipro research grant, Ramanujan Fellowship, the Infosys Centre for AI, IIIT Delhi, and ihub-Anubhuti-iiitd Foundation, set up under the NM-ICPS scheme of the Department of Science and Technology, India.
It is also part of the Tanbih mega-project, developed at the Qatar Computing Research Institute, HBKU, which aims to limit the impact of ``fake news,'' propaganda, and media bias by making users aware of what they are reading, thus promoting media literacy and critical thinking.

\bibliographystyle{acl_natbib}
\bibliography{anthology,custom}

\clearpage
\appendix

\section*{Appendix}
\label{sec:appendix}



\section{Implementation Details and Hyper-parameter Values} \label{sec:hyperparameters}

We trained all our models using PyTorch on NVIDIA Tesla V100 GPU, with 32 GB dedicated memory, CUDA-11.2 and cuDNN-8.1.1 installed. For the unimodal models, we imported all the pre-trained weights from the \texttt{TORCHVISION.MODELS}\footnote{\url{http://pytorch.org/docs/stable/torchvision/models.html}}, a sub-package of the PyTorch framework. We initialized the remaining weights randomly using a zero-mean Gaussian distribution with a standard deviation of 0.02. We train \model\ in a setup considering only \textit{harmful} class data from \olddata\ \cite{pramanick-etal-2021-momenta-multimodal}. We extended it by manually annotating for \textit{harmful} targets, followed by including \textit{non-harmful} examples using automated entity extraction (textual and visual) strategies for training/validation splits and manual annotation (for both harmful and non-harmful) for the test split. 

When training our models and exploring various values for the different model hyper-parameters, we experimented with using the Adam optimizer \cite{kingma2014adam} with a learning rate of $1e^{-4}$, a weight decay of $1e^{-5}$, and a Binary Cross-Entropy (BCE) loss as the objective function. We extensively fine-tuned our experimental setups based upon different architectural requirements to select the best hyper-parameter values. We also used early stopping for saving the best intermediate checkpoints. Table~\ref{tab:hyperparameters} gives more detail about the hyper-parameters we used for training. On average, it took approximately 2.5 hours to train a multi-modal neural model.

\begin{table}[h!]
\centering
\resizebox{\columnwidth}{!}
{
\begin{tabular}{c  l  c  c  c  c  c  c }
\hline
& &  \bf BS & \bf \#Epochs & \bf LR & \bf V-Enc & \bf T-Enc & \bf \#Param \\
\hline

\multirow{4}{*}{\centering \bf UM} & GRU & 32 & 25 & 0.0001 & - & \texttt{bert} & 2M\\
& XLNet & 16 & 20 & 0.0001 & - & \texttt{xlnet} & 116M\\
& VGG16 & 32 & 25 & 0.0001 & VGG16 & - & 117M\\
& ViT & 16 & 20 & 0.0001 & \texttt{vit} & - & 86M\\
\hline
\multirow{5}{*}{ \centering \bf MM} & MMFT & 16 & 20 & 0.001 & ResNet-152 & \texttt{bert} & 170M\\
& MMBT & 16 & 20 & 0.001 & ResNet-152 & \texttt{bert} & 169M\\
& ViLBERT* & 16 & 10 & 0.001 & Faster RCNN & \texttt{bert} & 112M\\
& V-BERT* & 16 & 10 & 0.001 & Faster RCNN & \texttt{bert} & 247M\\ \cline{2-8}
& \model & 16 & 30 & 0.0001 & \texttt{vit} & \texttt{bert} & 111M \\
\hline
\end{tabular}}
\caption{Hyperparameters summary. [BS$\rightarrow$Batch Size; LR$\rightarrow$Learning Rate; V/T-Enc$\rightarrow$Vision/Text-Encoder; \texttt{vit}$\rightarrow$\texttt{vit-base-patch16-224-in21k}; \texttt{bert}:$\rightarrow$\texttt{bert-base-uncased}; \texttt{xlnet}$\rightarrow$\texttt{xlnet-base-uncased}].}
\label{tab:hyperparameters}
\end{table}

\begin{table*}[htb!]
\centering
\resizebox{\textwidth}{!}{
\begin{tabular}{l|c|cc|cc|c|cc|cc|c|cc|cc}
\hline
\multicolumn{1}{c}{} & \multicolumn{5}{|c}{\textbf{Test Set A}} & \multicolumn{5}{|c}{\textbf{Test Set B}} & \multicolumn{5}{|c}{\textbf{Test Set C}} \\ \cline{2-16} 
\multicolumn{1}{c|}{} & \multicolumn{1}{c|}{} & \multicolumn{2}{c|}{\textbf{Not-harmful}} & \multicolumn{2}{c|}{\textbf{Harmful}} & \multicolumn{1}{c|}{} & \multicolumn{2}{c|}{\textbf{Not-harmful}} & \multicolumn{2}{c|}{\textbf{Harmful}} & \multicolumn{1}{c|}{} & \multicolumn{2}{c|}{\textbf{Not-harmful}} & \multicolumn{2}{c}{\textbf{Harmful}} \\ \cline{3-6}\cline{8-11}\cline{13-16} 
\multicolumn{1}{c|}{\multirow{-3}{*}{\textbf{Approach}}} &  \multicolumn{1}{c|}{\multirow{-2}{*}{\textbf{F1}}} & \multicolumn{1}{c|}{\textbf{P}} & \multicolumn{1}{c|}{\textbf{R}} & \multicolumn{1}{c|}{\textbf{P}} & \multicolumn{1}{c|}{\textbf{R}} & \multicolumn{1}{c|}{\multirow{-2}{*}{\textbf{F1}}} & \multicolumn{1}{c|}{\textbf{P}} & \multicolumn{1}{c|}{\textbf{R}} & \multicolumn{1}{c|}{\textbf{P}} & \multicolumn{1}{c|}{\textbf{R}} & \multicolumn{1}{c|}{\multirow{-2}{*}{\textbf{F1}}} & \multicolumn{1}{c|}{\textbf{P}} & \multicolumn{1}{c|}{\textbf{R}} & \multicolumn{1}{c|}{\textbf{P}} & \multicolumn{1}{c}{\textbf{R}}\\ \hline
CE & 0.7411 & \multicolumn{1}{l|}{0.71} & 0.78 & \multicolumn{1}{l|}{0.77} & 0.71 & 0.4847 & \multicolumn{1}{c|}{0.78} & \textbf{0.95} & \multicolumn{1}{c|}{0.29} & 0.07 & 0.4829 & \multicolumn{1}{c|}{0.83} & \textbf{0.93} & \multicolumn{1}{c|}{0.17} & 0.06 \\ \hline 
\embharm\ & 0.7250 & \multicolumn{1}{l|}{0.75} & 0.66 & \multicolumn{1}{l|}{0.71} & 0.79 & 0.5544 & \multicolumn{1}{c|}{0.81} & 0.72 & \multicolumn{1}{c|}{0.3} & 0.41 & 0.5658 & \multicolumn{1}{c|}{0.88} & 0.68 & \multicolumn{1}{c|}{0.27} & 0.56 \\ \hline 
CI & 0.7729 & \multicolumn{1}{l|}{0.74} & 0.82 & \multicolumn{1}{l|}{0.81} & 0.73 & 0.5174 & \multicolumn{1}{c|}{0.79} & 0.89 & \multicolumn{1}{c|}{0.29} & 0.15 & 0.5314 & \multicolumn{1}{c|}{0.84} & 0.87 & \multicolumn{1}{c|}{0.23} & 0.19 \\ \hline 
\content\ + \embharm\ & 0.7406 & \multicolumn{1}{l|}{0.71} & 0.78 & \multicolumn{1}{l|}{0.78} & 0.7 & 0.5775 & \multicolumn{1}{c|}{0.82} & 0.74 & \multicolumn{1}{c|}{0.33} & 0.44 & 0.5840 & \multicolumn{1}{c|}{\textbf{0.89}} & 0.7 & \multicolumn{1}{c|}{0.29} & 0.57 \\ \hline 
\content\ + \contimg\ (concat) & 0.7361 & \multicolumn{1}{l|}{0.71} & 0.77 & \multicolumn{1}{l|}{0.77} & 0.7 & 0.4230 & \multicolumn{1}{c|}{0.74} & 0.51 & \multicolumn{1}{c|}{0.18} & 0.37 & 0.4125 & \multicolumn{1}{c|}{0.82} & 0.43 & \multicolumn{1}{c|}{0.17} & 0.56 \\ \hline 
\content\ + \contimg\ (\mmlrbp) & 0.7790 & \multicolumn{1}{l|}{0.74} & \textbf{0.84} & \multicolumn{1}{l|}{\textbf{0.83}} & 0.72 & 0.5079 & \multicolumn{1}{c|}{0.81} & 0.57 & \multicolumn{1}{c|}{0.26} & 0.52 & 0.4857 & \multicolumn{1}{c|}{0.88} & 0.5 & \multicolumn{1}{c|}{0.22} & 0.69 \\ \hline
\embharm\ + \contimg\ (concat) & 0.6609 & \multicolumn{1}{l|}{0.67} & 0.6 & \multicolumn{1}{l|}{0.66} & 0.72 & 0.4964 & \multicolumn{1}{c|}{0.78} & 0.65 & \multicolumn{1}{c|}{0.23} & 0.37 & 0.4421 & \multicolumn{1}{c|}{0.81} & 0.57 & \multicolumn{1}{c|}{0.15} & 0.38 \\ \hline 
\embharm\ + \contimg\ (\mmlrbp) & 0.7260 & \multicolumn{1}{l|}{0.74} & 0.67 & \multicolumn{1}{l|}{0.72} & 0.78 & 0.5470 & \multicolumn{1}{c|}{0.8} & 0.74 & \multicolumn{1}{c|}{0.29} & 0.37 & 0.4836 & \multicolumn{1}{c|}{0.83} & 0.68 & \multicolumn{1}{c|}{0.17} & 0.31\\ \hline 
{\color[HTML]{000000} \model} & {\color[HTML]{000000} \textbf{0.7845}} & \multicolumn{1}{l|}{{\color[HTML]{000000} 0.74}} & {\color[HTML]{000000} 0.81} & \multicolumn{1}{l|}{{\color[HTML]{000000} 0.74}} & {\color[HTML]{000000} \textbf{0.86}} & {\color[HTML]{000000} \textbf{0.6498}} & \multicolumn{1}{c|}{{\color[HTML]{000000} \textbf{0.83}}} & {\color[HTML]{000000} 0.79} & \multicolumn{1}{c|}{{\color[HTML]{000000} \textbf{0.38}}} & {\color[HTML]{000000} \textbf{0.69}} & {\color[HTML]{000000} \textbf{0.6412}} & \multicolumn{1}{c|}{{\color[HTML]{000000} 0.86}} & {\color[HTML]{000000} 0.76} & \multicolumn{1}{c|}{{\color[HTML]{000000} \textbf{0.36}}} & {\color[HTML]{000000} \textbf{0.7}} \\ \hline
\end{tabular}}
\caption{Ablation results for \model\ and its variants for Test Sets A, B, and C.}
\label{tab:AAbRes}
\end{table*}


\section{Ablation Study}
\label{sec:ablation}
In this section, we present some ablation studies for sub-modules of \model\ based on \content, \embharm, \conttxt, and \contimg, examined in isolation and in combinations, and finally for \model\ using \contmm.

\subsection{Test Set A}
As observed in the comparisons made with the other baseline systems for the Test Set A in Table \ref{tab:ARes}, the overall range of the F1 scores is relatively higher with the lowest value being 0.66 for XLNet (text-only) model. The results for unimodal systems, as can be observed in Table~\ref{tab:AAbRes}, is satisfactory with values of 0.74, 0.73, and 0.77 for \content\, \embharm,\ and \contimg\ unimodal systems, respectively. For multimodal systems, we can observe distinct lead for the \mmlrbp-based fusion strategy, for both \content\ and \embharm\ systems over the concatenation-based approach, except for \embharm's recall drop by 7\%.
Finally \model\ yields the best overall F1 score of 0.78.

\subsection{Test Set B}
With \textit{context} not having any harmfulness cues for a given meme when considered in isolation, the unimodal \content\ module performs the worst with 0.48 F1 score, and 0.07 recall for the \textit{harmful} class, in the open-ended setting of Test Set B. In contrast, \embharm\ yields an impressive F1 score of 0.55, and a \textit{harmful} recall of 0.41. This relative gain of 7\% in terms of F1 score could be due to the presence of explicit harmfulness cues. The complementary effect of considering contextual information can be inferred from the joint modeling of \content\ and \embharm,\ to obtained \conttxt,\ that enhances the F1 score and the \textit{harmful} recall by 2\% and 3\%, respectively (see Table~\ref{tab:AAbRes}). Unimodal assessment of \contimg\ performs moderately with an F1 score of 0.51, but with a poor \textit{harmful} recall of 0.15. \mmlrbp,\ towards joint-modeling of \content\ and \contimg\ yields a significant boost in the \textit{harmful} recall to 0.52 (see Table~\ref{tab:AAbRes}). On the other hand, \mmlrbp-based fusion of \embharm\ and \contimg\ yields 0.54 F1 score, which is 1\% below that for the unimodal \embharm\ system. This emphasizes the importance of accurately modeling the embedded harmfulness, besides \emph{augmenting} with additional context. A complementary impact of \content, \embharm, and \contimg\ is observed for \model\ with a balanced F1 score of 0.6 and a competitive \textit{harmful} recall value of 0.69.  

\subsection{Test Set C}
As observed in the previous scenario, the unimodal models for \content\ yield a low F1 score of 0.48 and the worst \textit{harmful} recall value of 0.06. Much better performance is observed for unimodal setups including \embharm, and its joint modelling with \content\ with improved F1 scores of 0.56 and 0.58, respectively, along with the \textit{harmful} recall score of 0.56 and 0.57, respectively. \contimg\ based unimodal evaluation again yields a moderate F1 score of 0.53 (see Table~\ref{tab:AAbRes}), along with a poor \textit{harmful} recall of 0.19, which shows its inadequacy to model harmful targeting on its own. For multimodal setups, the joint modelling of \content\ and \contimg\ benefits from \mmlrbp\ based fusion, yielding a gain of 7\% and 13\% in terms of F1 score and \textit{harmful} recall, respectively. This confirms the importance of contextual multimodal semantic alignment. Correspondingly, joint multimodal modelling of \embharm\ and \contimg\  regresses the unimodal affinity within the \embharm.\ Finally, \model\ outperforms all other systems in this category with the best F1 score of 0.64, with a decent \textit{harmful} recall score of 0.7. 

The experimental results here are for comparison and analysis of the optimal set of design and baseline choices. We should note that we performed extensive experiments as part of our preliminary investigation, with different contextual modelling strategies, attention mechanisms, modelling choices, etc., to reach a conclusive architectural configuration that show promise for addressing the task of target detection in harmful memes.

\section{Error Analysis}
\label{sec:err_analysis}

\begin{figure}[!t]
    \begin{minipage}{0.62\columnwidth}%
    \begin{subfigure}{\textwidth}%
        \includegraphics[width=\linewidth]{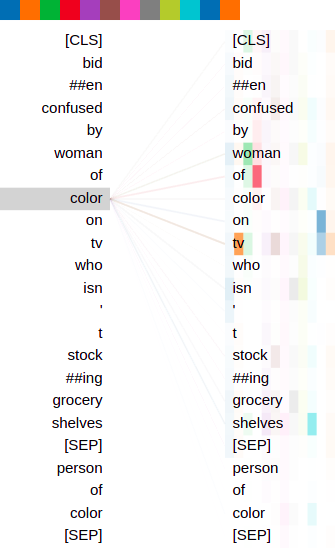}
        \caption{L-AT}
        \label{fig:err_text}
    \end{subfigure}
    \end{minipage}%
    \hspace{0.2mm}
    \begin{minipage}{0.29\columnwidth}%
        \begin{minipage}{\textwidth}%
            \begin{subfigure}{\linewidth}%
                \includegraphics[width=\linewidth]{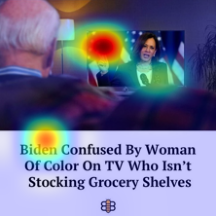}
                \caption{MM-AT-CLIP} 
                \label{fig:err_clip}
            \end{subfigure}
        \end{minipage}\\
        \begin{minipage}{\textwidth}%
            \begin{subfigure}{\linewidth}%
                \includegraphics[width=\linewidth]{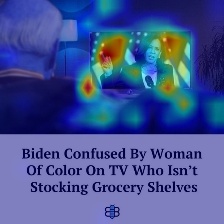}
                \caption{V-AT-\model} 
                \label{fig:err_disarm}
            \end{subfigure}
        \end{minipage}\\
        \begin{minipage}{\textwidth}%
            \begin{subfigure}{\linewidth}%
                \includegraphics[width=\linewidth]{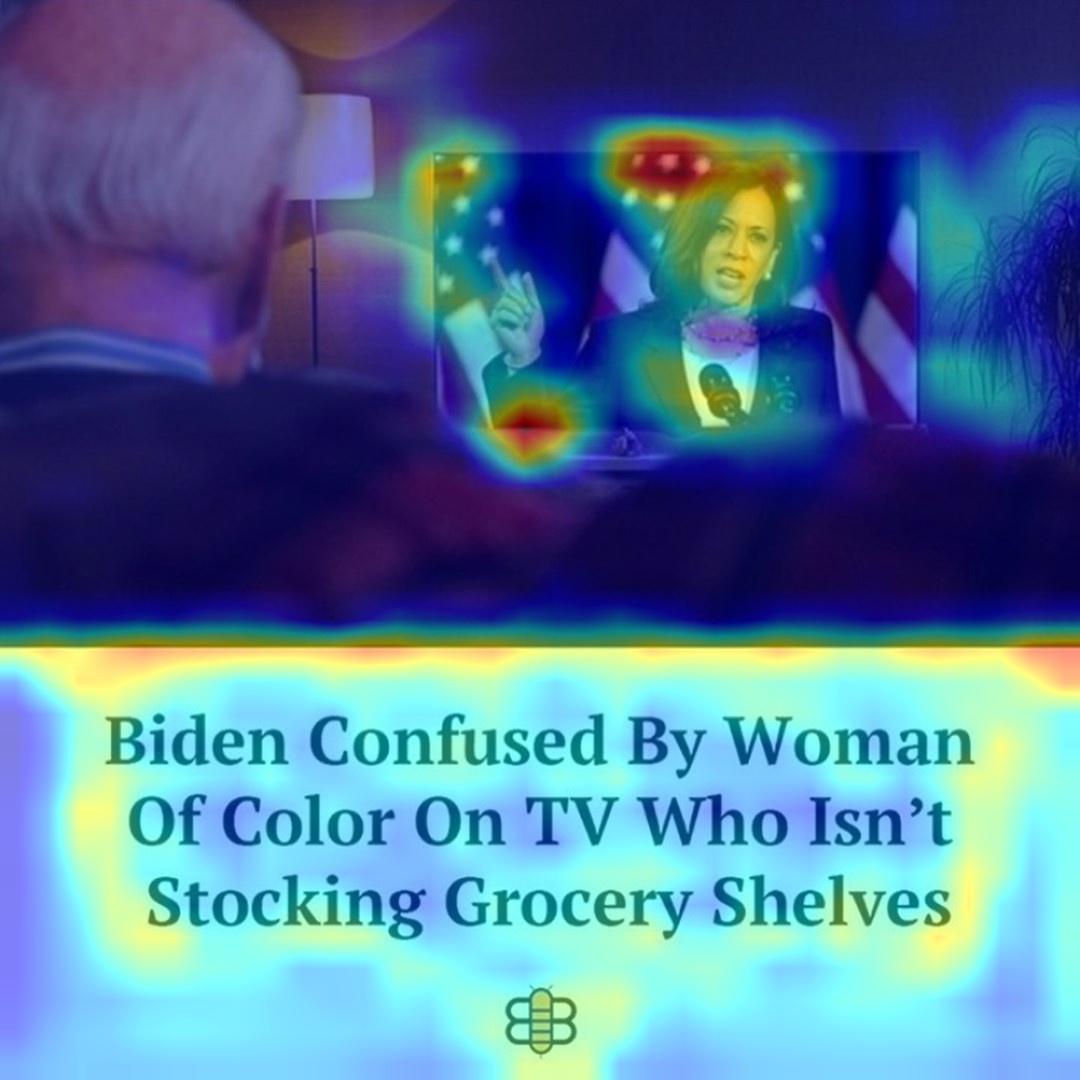}
                \caption{V-AT-ViLBERT} 
                \label{fig:err_vilbert}
            \end{subfigure}
        \end{minipage}
        
    \end{minipage}\\\vspace{0.4mm}
\subfloat{\framebox[\columnwidth]{\parbox{\dimexpr\linewidth-2\fboxsep-2\fboxrule}{Target Candidate$\rightarrow$\textcolor{blue}{\texttt{person of color}} \justify{Context$\rightarrow$\textcolor{blue}{\texttt{During the evening of the VP debates, Joe Biden settled down on his soft couch with a glass of warm milk to watch this.}}}}}}
\caption{Comparison of attention maps for miclassification between \model\ [(a), (b) \& (c)] and ViLBERT [(d)] using BertViz and Grad-CAM.}
\label{fig:err_analysis}
\end{figure}

\begin{figure*}[t]
\centering
\subfloat[\label{fig:analogy_ex}Harmful analogy]{
\includegraphics[width=0.185\textwidth, height=0.185\textwidth]{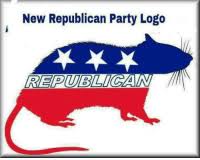}}\hspace{0.2em}
\subfloat[\label{fig:graphic_ex}Sensitive visuals]{
\includegraphics[width=0.185\textwidth, height=0.185\textwidth]{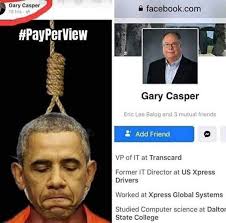}}\hspace{0.2em}
\subfloat[\label{fig:barackgay_ex}Political grounds]{
\includegraphics[width=0.185\textwidth, height=0.185\textwidth]{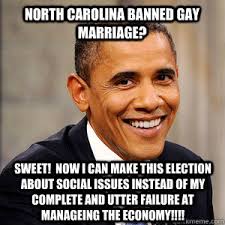}}\hspace{0.2em}
\subfloat[\label{fig:bushterr_ex}Religious grounds]{
\includegraphics[width=0.185\textwidth, height=0.185\textwidth]{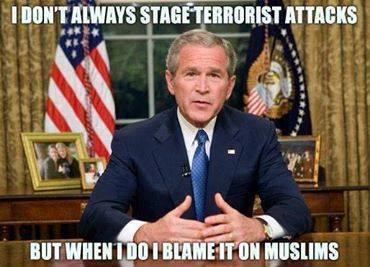}}\hspace{0.2em}
\subfloat[\label{fig:baracknation_ex}International threat]{
\includegraphics[width=0.185\textwidth, height=0.185\textwidth]{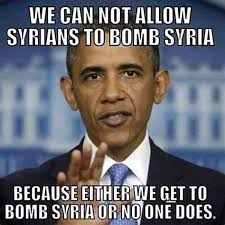}}\hspace{0.2em}
\caption{Examples of memes depicting different types (a)--(e) of \textit{harmful} targeting. }
\label{fig:meme:harmtar_examples}
\end{figure*}

It is evident from the results shown in Tables \ref{tab:ARes} and \ref{tab:CRes} that \model\ still has  shortcomings. Examples like the one shown in Fig.~\ref{fig:err_analysis} are seemingly \textit{harmless}, both textually and visually, but imply serious \textit{harm} to a \textit{person of color} in an implicit way.

This kind of complexity can be challenging to model without providing additional context about the meme like \textit{people of colour face racial discrimination all over the world}. This is also analogous to a fundamental challenge associated with detecting implicit hate \cite{MacAvaney2019Hate}. In this particular example, despite modelling contextual information explicitly in \model,\ it misclassifies this meme anyway.

Even though the context obtained for this meme pertains to its content (see Fig.~\ref{fig:err_analysis}), it does not relate to \emph{global racial prejudice}, which is key to ascertaining it as a harmfully targeting meme. Moreover, besides context, visuals and the message embedded within the meme do not convey definite harm when considered in isolation. This error can be inferred clearly from the embedded-harmfulness, contextualised-visuals, and the visuals being attended by \model\ as depicted in Fig.~\ref{fig:err_text}, Fig.~\ref{fig:err_clip}, and Fig.~\ref{fig:err_disarm}, respectively. On the other hand, as shown in the visual attention plot for ViLBERT in Fig.~\ref{fig:err_vilbert}, the field of view that is being attended encompasses the visuals of \textit{Kamala Harris}, who is the \textit{person of colour} that i sbeing primarily targeted by the meme. 
Besides the distinct attention on the primary target-candidate within the meme, ViLBERT could have leveraged the pre-training it received from Conceptual Captions (CC) \cite{sharma2018conceptual}, a dataset known for its diverse coverage of complex textual descriptions. This essentially highlights the importance of making use of multimodal pre-training using the dataset that is not as generic as MS COCO \cite{lin2014microsoft}, but facilitates modelling of the complex real-world multimodal information, especially for tasks related to memes.

\begin{table*}[tbh]
\centering
\resizebox{\textwidth}{!}{%
\begin{tabular}{c|c|c|c|c|c}
\hline
\multicolumn{3}{c|}{\textbf{Harmful meme}}                                                                  & \multicolumn{3}{c}{\textbf{Not-harmful meme}}                                     \\ \hline
\textbf{Individual}  & \textbf{Organization}                          & \textbf{Community}                   & \textbf{Individual}           & \textbf{Organization}     & \textbf{Community}      \\ \hline
joe biden (333)      & democratic party (184) & mexicans (11) & donald trump (106)            & green party (189)         & trump supporters (86)   \\ \hline
donald trump (285)   & republican party (130)                         & black (7)    & republican voter (102)        & biden camp (162)          & white (50)              \\ \hline
barack obama (142)   & libertarian party (44)                         & muslim (7)                           & barack obama (94)             & communist party (114)     & african american (47)   \\ \hline
hillary clinton (35) & cnn (6)                                        &islam (6)    & joe biden (47)                & america (64)              & democrat officials (45) \\ \hline
mike pence (13)      & government (5)                                 & russian (5)                          & alexandria ocasio cortez (44) & trump administration (52) & republican (44)         \\ \hline
\end{tabular}
}
\caption{The top-5 most frequently referenced entities in each harmfulness class and their target categories. The total frequency for each word is shown in parentheses.}
\label{tab:lexsum}
\end{table*}

\section{Annotation Guidelines}
\label{sec:annotations}

Before discussing some details about the annotation process, revisiting the definition of \textit{harmful} memes would set the pretext towards consideration of \textit{harmful} targeting and \textit{non-harmful} referencing. According to \citet{pramanick-etal-2021-momenta-multimodal}, a harm can be expressed as an abuse, an offence, a disrespect, an insult, or an insinuation of a targeted entity or any socio-cultural or political ideology, belief, principle, or doctrine associated with that entity.
The harm can also be in the form of a more subtle attack such as mocking or ridiculing a person or an idea.

Another common understanding\footnote{\url{https://reportharmfulcontent.com/advice/other/further-advice/harmful-content-online-an-explainer}}\footnote{\url{https://swgfl.org.uk/services/report-harmful-content}}\footnote{\url{https://saferinternet.org.uk/report-harmful-content}} about the harmful content is that it could be anything online that causes distress. It is an extremely subjective phenomenon, wherein what maybe be harmful to some might not be considered an issue by others. This makes it significantly challenging to characterize and hence to study it via the computational lens. 

Based on a survey of 52 participants, \citet{harm_framework} defines online harm to be any violating content that results in any (or a combination) of the following four categories: (\emph{i})~physical harm, (\emph{ii})~emotional harm, (\emph{iii})~relational harm, and (\emph{iv})~financial harm. 
With this in mind, we define two types of referencing that we have investigated in our work within the context of internet memes: (\emph{i})~\textit{harmful} and (\emph{ii})~\textit{non-harmful}.

\subsection{Reference Types}

\paragraph{Harmful.} The understanding about harmful referencing (\textit{targeting}) in memes, can be sourced back to the definition of harmful memes by \citet{pramanick-etal-2021-momenta-multimodal}, wherein a social entity is subjected to some form of ill-treatment such as mental abuse, psycho-physiological injury, proprietary damage, emotional disturbance, or public image damage, based on their background (bias, social background, educational background, etc.) by a meme author.
\paragraph{Not-harmful.} Non-harmful referencing in memes is any benign mention (or depiction) of a social entity via humour, limerick, harmless pun or any content that does not cause distress. Any reference that is \textit{not} harmful falls under this category.

\subsection{Characteristics of Harmful Targeting}

There are several factors that collectively facilitate the characterisation of \textit{harmful} targeting in memes. Here are some:
\begin{enumerate}
    \item A prominent way of harmfully targeting an entity in a meme is by leveraging sarcastically harmful analogies, framed via either textual or visual instruments (see Fig.~\ref{fig:analogy_ex}).
    \item There could be multiple entities being harmfully targeted within a meme as depicted in Fig.~\ref{fig:annot_ex}. Hence, annotators were asked to provide all such targets as harmful, with no exceptions.
    \item A harmful targeting within a meme could have visual depictions that are either gory, violent, graphically sensitive, or pornographic (see Fig.~\ref{fig:graphic_ex}).
    \item Any meme that insinuates an entity on either social, political, professional, religious grounds, can cause harm (see Fig.~\ref{fig:barackgay_ex} and \ref{fig:bushterr_ex}).
    \item Any meme that implies an explicit/implicit threat to an individual, a community, a national or an international entity is harmful (see Fig.~\ref{fig:bushterr_ex} and \ref{fig:baracknation_ex}). 
    \item Whenever there is any ambiguity regarding the harmfulness of any reference being made, we requested the annotators to proceed following the best of their understanding.
\end{enumerate}
\section{\extharmp\ Characteristics}

Below, we perform some analysis of the lexical content of the length of the meme text.

\subsection{Lexical Analysis}

\begin{figure*}[t!]
\centering
\subfloat[{\centering Trump}\label{fig:lentrump}]{
\includegraphics[width=0.30\textwidth, height=0.18\textwidth]{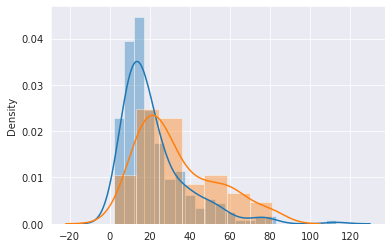}}\hspace{0.1em}
\subfloat[{Republican Party}\label{fig:lenrepublican}]{
\includegraphics[width=0.30\textwidth, height=0.18\textwidth]{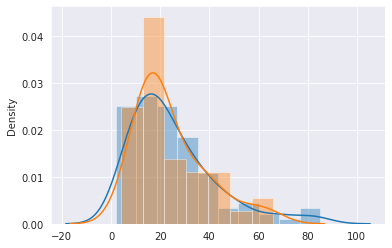}}\hspace{0.1em}
\subfloat[{\centering Mexican}\label{fig:lenmexican}]{
\includegraphics[width=0.30\textwidth, height=0.18\textwidth]{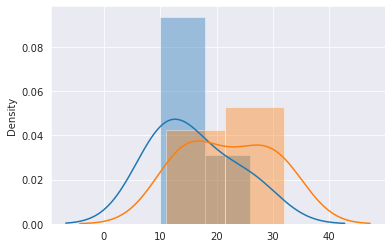}} \\
\subfloat[{\centering Biden}\label{fig:lenbiden}]{
\includegraphics[width=0.30\textwidth, height=0.18\textwidth]{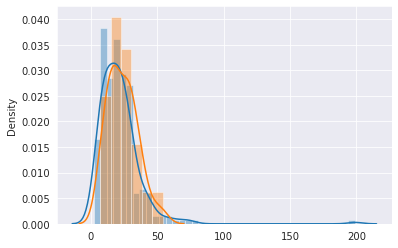}}\hspace{0.1em}
\subfloat[{Democratic Party}\label{fig:democraticparty}]{
\includegraphics[width=0.30\textwidth, height=0.18\textwidth]{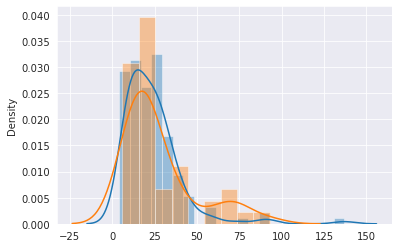}}\hspace{0.1em}
\subfloat[{\centering Black}\label{fig:lenblack}]{
\includegraphics[width=0.30\textwidth, height=0.18\textwidth]{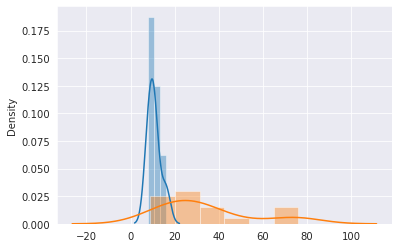}} 
\caption{Distributions of the OCR's length for the memes of top-5 harmful references: harmful (Blue) and non-harmful (Orange). The depiction is for Individual: (a) and (d); Organization: (b) and (e); and Community: (c) and (f).}
\label{fig:meme-text_lengths}
\end{figure*}

Interestingly, a significant number of memes are disseminated making references to popular \textit{individuals} such as \textit{Joe Biden, Donald Trump, etc.}, as can be observed for individual sub-categories (for both harmful and non-harmful memes) in Table~\ref{tab:lexsum}.

We can see in Table \ref{tab:lexsum} that for \textit{harmful}--\textit{organization}, the top-5 harmfully targeted organizations include the top-2 leading political organizations in the USA (the \emph{Democratic Party} and the \emph{Republican Party}), which are of significant political relevance, followed by the \emph{Libertarian Party}, a media outlet (\emph{CNN}), and finally the generic \emph{government}. At the same time, non-harmfully referenced organizations includes the \emph{Biden camp} and the \textit{Trump administration}, which are mostly leveraged for harmfully targeting (or otherwise) the associated public figure. Finally, communities such as \emph{Mexicans, Black, Muslim, Islam}, and \emph{Russian} are often immensely prejudiced against online, and thus also in our meme dataset. At the same time, non-harmfully targeted communities such as the \emph{Trump supporters} and the \emph{African Americans} are not targeted as often as the aforementioned ones, as we can see in Table~\ref{tab:lexsum}. 

The above analysis of the lexical content of the memes in our datasets largely emphasizes the inherent bias that multimodal content such as memes can exhibit, which in turn can have direct influence on the efficacy of machine/deep learning-based systems for detecting the entities targeted by harmful memes. The reasons for this bias are mostly linked to societal behaviour at the organic level, and the limitations posed by current techniques to process such data. The mutual exclusion for harmful vs. non-harmful categories for community shows the inherent bias that could pose a challenge, even for the best multi-modal deep neural systems. The high pervasiveness of a few prominent keywords could effectively lead to increasing bias towards them for specific cases.  At the same time, the significant overlap observed in Table~\ref{tab:lexsum} for the enlisted entities, between harmful and not-harmful individuals, highlights the need for sophisticated multi-modal systems that can effectively reason towards making a complex decision like detecting harmful targeting within memes, rather than exploit the biases towards certain entities in the training data. 

\subsection{Meme-Message Length Analysis}

Most of the \textit{harmful} memes are observed to be created using texts of length 16--18 (see Fig.~\ref{fig:meme-text_lengths}). At the same time, \textit{not-harmful} meme-text lengths have a relatively higher standard deviation, possibly due to the diversity of \textit{non-harmful} messages. \emph{Trump} and the \emph{Republic Party} have meme-text length distributions similar to the \textit{non-harmful} category: skewing left, but gradually decreasing towards the right. This suggests a varying content generation pattern amongst meme creators (see Fig.~\ref{fig:meme-text_lengths}). The meme-text length distribution for \emph{Biden} closely approximates a normal distribution with a low standard deviation. Both categories would pre-dominantly entail creating memes with shorter text lengths, possibly due to the popularity of \emph{Biden} amongst humorous content creators. A similar trend could be seen for the \emph{Democratic Party} as well, where most of the instances fall within the 50--75 meme-text length range. The overall harmful and non-harmful meme-text length distribution is observed to be fairly distributed across different meme-text lengths for \textit{Mexican}. At the same time, the amount of harm intended towards the \textit{Black} community is observed to be significantly higher, as compared to moderately distributed \textit{non-harmful} memes depicted by the corresponding meme-text length distribution in Fig.~\ref{fig:meme-text_lengths}.
\end{document}